%% file: main.tex
\documentclass[10pt,conference]{IEEEtran}
\usepackage{cite}
\usepackage{amsmath,amssymb,amsfonts}
\usepackage{algorithmic}
\usepackage{graphicx}
\usepackage{textcomp}
\usepackage{xcolor}
\usepackage[hyphens]{url}

\def\BibTeX{{\rm B\kern-.05em{\sc i\kern-.025em b}\kern-.08em
    T\kern-.1667em\lower.7ex\hbox{E}\kern-.125emX}}

\newcommand{\hrhpca}[1]{\textcolor{black}{#1}}
\newcommand{\hry}[1]{\textcolor{black}{#1}}
\newcommand{\hr}[1]{\textcolor{black}{#1}}

\newcommand{\shh}[1]{\textcolor{black}{#1}}
\usepackage{enumitem}
\setlist[itemize]{leftmargin=4mm}
\usepackage{amsmath}
\usepackage{amsfonts}
\usepackage{wrapfig}
\usepackage{booktabs} 
\usepackage{pifont}

\usepackage[normalem]{ulem}
\usepackage{multirow}
\usepackage[ruled,vlined,linesnumbered]{algorithm2e}
\SetKwComment{Comment}{$\triangleright$\ }{}
\SetKwComment{CommentL}{$\/\/$}{}
\SetKwFunction{FMain}{Main}
\usepackage{tikz}
\usetikzlibrary{fit,calc}

\newcommand{\tabincell}[2]{\begin{tabular}{@{}#1@{}}#2\end{tabular}} 
\usepackage{amssymb}
\usepackage{pifont}
\newcommand{\cmark}{\ding{51}}%
\newcommand{\xmark}{\ding{55}}%

\newcommand*{\tikzmka}[1]{\tikz[remember picture,overlay,] \node (#1) {};\ignorespaces}
\newcommand{\boxita}[1]{\tikz[remember picture,overlay]{\node[yshift=3pt,fill=#1,opacity=.15,fit={($(A)+(0.005\linewidth,0.2\baselineskip)$)($(B)+(0.02\linewidth,-.2\baselineskip)$)}] {};}\ignorespaces}

\newcommand*{\tikzmkc}[1]{\tikz[remember picture,overlay,] \node (#1) {};\ignorespaces}
\newcommand{\boxitc}[1]{\tikz[remember picture,overlay]{\node[yshift=3pt,fill=#1,opacity=.1,fit={($(A)+(0.005\linewidth,0.2\baselineskip)$)($(B)+(0.02\linewidth,-0.2\baselineskip)$)}] {};}\ignorespaces}
\colorlet{pink}{red!40}
\colorlet{cyan}{cyan!60}
\colorlet{orange}{orange!80}

\title{ViTCoD: Vision Transformer Acceleration via Dedicated Algorithm and Accelerator Co-Design} 

\author{
\IEEEauthorblockN{Haoran You\IEEEauthorrefmark{1},
Zhanyi Sun\IEEEauthorrefmark{2},
Huihong Shi\IEEEauthorrefmark{1}, 
Zhongzhi Yu\IEEEauthorrefmark{1},
Yang Zhao\IEEEauthorrefmark{2},\\
Yongan Zhang\IEEEauthorrefmark{1},
Chaojian Li\IEEEauthorrefmark{1},
Baopu Li\IEEEauthorrefmark{3} and
Yingyan Lin\IEEEauthorrefmark{1}\\}
\IEEEauthorblockA{\IEEEauthorrefmark{1}
Georgia Institute of Technology,
Atlanta, GA\\}
\IEEEauthorblockA{\IEEEauthorrefmark{2}Rice University, Houston, TX \quad \IEEEauthorrefmark{3}Oracle Health and AI, Redwood, CA}
\{hyou37, eiclab, zyu401, yzhang919, cli851, celine.lin\}@gatech.edu,
\{zs19, zy34\}@rice.edu, baopu.li@oracle.com
}

\begin{document}
\maketitle
\thispagestyle{plain}
\pagestyle{plain}


\begin{abstract}

Vision Transformers (ViTs) have achieved state-of-the-art performance on various vision tasks. However, ViTs' self-attention module is still arguably a major bottleneck, limiting their achievable hardware efficiency and more extensive applications to resource constrained platforms. Meanwhile, existing accelerators dedicated to NLP Transformers are not optimal for ViTs. This is because there is a large difference between ViTs and Transformers for natural language processing (NLP) tasks: ViTs have a relatively fixed number of input tokens, whose attention maps can be pruned by up to 90\% even with fixed sparse patterns, \hry{without severely hurting the model accuracy (e.g., $<=$1.5\% under 90\% pruning ratio)}; while NLP Transformers need to handle input sequences of varying numbers of tokens and rely on on-the-fly predictions of dynamic sparse attention patterns for each input to achieve a decent sparsity (e.g., $>=$50\%). To this end, we propose a dedicated algorithm and accelerator co-design framework dubbed \textbf{ViTCoD} for accelerating ViTs. Specifically, 
\uline{on the algorithm level}, ViTCoD prunes and polarizes the attention maps to have either denser or sparser fixed patterns for regularizing two levels of workloads without hurting the accuracy, largely reducing the attention computations while leaving room for alleviating the remaining dominant data movements; on top of that, we further integrate a lightweight and learnable auto-encoder module to enable trading the dominant high-cost data movements for lower-cost computations.
\uline{On the hardware level}, we develop a dedicated accelerator to simultaneously coordinate the aforementioned enforced denser and sparser workloads for boosted hardware utilization, while integrating on-chip encoder and decoder engines to leverage ViTCoD's algorithm pipeline for much reduced data movements.
Extensive experiments and ablation studies validate that ViTCoD largely reduces the dominant data movement costs, achieving speedups of \hry{up to} \hr{235.3$\times$, 142.9$\times$, 86.0$\times$, 10.1$\times$, and 6.8$\times$} over general computing platforms CPUs, EdgeGPUs, GPUs, and prior-art Transformer accelerators SpAtten and Sanger \hry{under an attention sparsity of 90\%}, respectively.
Our code implementation is available at \textcolor[RGB]{227,23,13}{\url{https://github.com/GATECH-EIC/ViTCoD}}.

\end{abstract}



\input{sections/1-Introduction}

\input{sections/2-Related_Work}

\input{sections/3-Methods}

\input{sections/4-Experiments}
\input{sections/5-Conclusion}

\section*{Acknowledgement}
We would like to acknowledge the funding support from NSF EPCN program (Award number: 1934767) and RTML (Award number: 1937592) for this project.

\bibliographystyle{IEEEtranS}
\bibliography{refs}

\end{document}

%% file: sections/1-Introduction.tex
\section{Introduction}
\label{sec:intro}








We have recently witnessed the amazing success and increasing interest of developing attention-based Transformer architectures for both natural language processing (NLP) and computer vision (CV) tasks.
The powerful performance of Transformers largely benefits from their self-attention module that is capable of extracting global context information \cite{transformer,lite_transformer,vit}.
However, the self-attention module comes at a cost of inefficiency during both training and inference due to its quadratic complexity dependency on the number of input tokens, and has been recognized as a major efficiency bottleneck for the inference acceleration of Transformers. 
For example, the self-attention module of the GPT-2 model \cite{radford2019language} accounts for over 50\% of the total latency measured on a TITAN Xp GPU \cite{wang2021spatten}; 
This percentage increases to 69\% for LeViT-128~\cite{graham2021levit} when measured on an EdgeGPU \cite{edgegpu}. 
To alleviate the bottleneck complexity of self-attention modules, sparse attention techniques have emerged as a promising solution and been considered by both algorithm \cite{zaheer2020big,beltagy2020longformer,treviso2021predicting} and hardware acceleration \cite{ham2021elsa,wang2021spatten,lu2021sanger,qu2022dota} works.

\begin{figure}[t]
    \centering
    \includegraphics[width=0.96\linewidth]{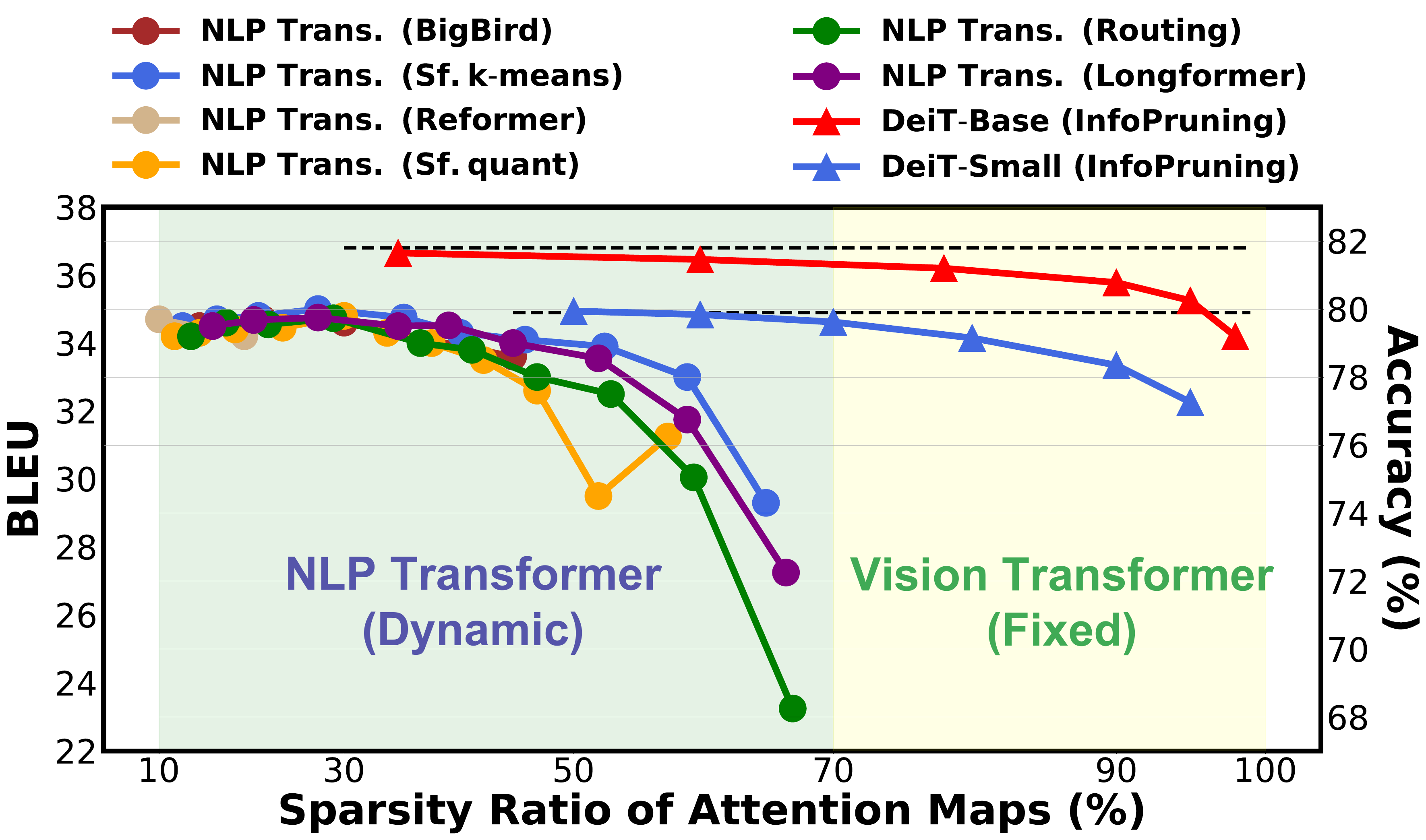}
    \vspace{-0.5em}
    \caption{Comparison between NLP Transformers and ViTs in terms of BLEU-sparsity or accuracy-sparsity trade-offs. Note that for NLP Transformer, we collect the results on machine translation task, IWSLT EN $\rightarrow$ DE, following \cite{treviso2021predicting}; For ViTs, we apply an info-based pruning technique on DeiT-Base/Small models and classification task (e.g., ImageNet), following \cite{kim2021rethinking}.
    }
    \vspace{-1.5em}
    \label{fig:comp_vit_nlp}
\end{figure}

\hry{
Despite their great promise, existing sparse attention accelerators or algorithm-accelerator co-design works (e.g., Sanger \cite{lu2021sanger}) focus on accelerating NLP Transformers, and adopt hardware designs with on-the-fly sparse attention prediction and high reconfigurability in order to handle the varying number of input tokens in NLP. As such, those techniques are not optimal for accelerating Vision Transformers (ViTs), which feature stark differences from NLP Transformers. \textbf{Next, we discuss the differences and corresponding new opportunities or challenges for efficient acceleration of ViTs}:
\uline{First}, ViTs have a relatively fixed number of input tokens during both training and inference (e.g., a commonly adopted token size of $16\times16$ for an image resolution of $224\times224$, which leads to a total of 196 tokens), while NLP Transformers adopt input-dependant varying numbers of tokens across different NLP datasets/tasks. 
ViTs' relatively fixed number of tokens offers an opportunity to design ViT accelerators, which can potentially avoid on-the-fly sparse attention pattern prediction adapting to each input, via co-designing with sparse ViT algorithms.
\uline{Second}, as shown in Fig. \ref{fig:comp_vit_nlp}, ViTs allow their attention maps to be pruned by up to 90\%$\sim$95\% with \textit{fixed} sparse patterns for all inputs without significant accuracy drops, whereas NLP Transformers often can only allow a medium level of sparsity ratio (e.g., 50\% $\sim$ 70\%) even for \textit{dynamic} sparse attention patterns \cite{treviso2021predicting} if aiming for small/negligible accuracy drops.}

\begin{figure}[t]
    \centering
    \includegraphics[width=\linewidth]{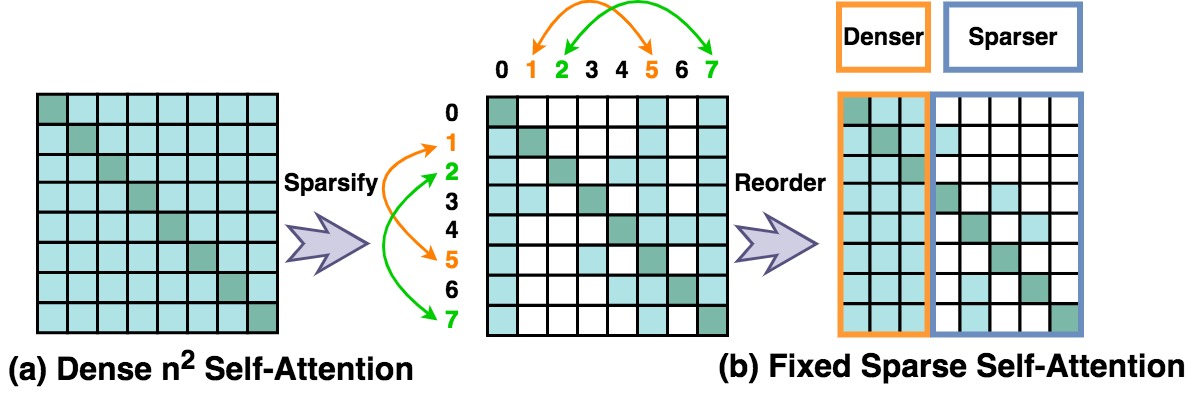}
      \vspace{-2.em}
    \caption{\hry{Illustrating the fixed sparse attention mask.}}
    \label{fig:split_and_conquer}
    \vspace{-1.5em}
\end{figure}

The aforementioned differences bring about both \textbf{new opportunities and challenges} for accelerating ViTs. 
\hry{On one hand, the fixed sparse patterns in ViTs can alleviate the stringent need for adopting on-the-fly sparse attention pattern prediction and highly reconfigurable processing element (PE) designs.}
On the other hand, ViTs' allowed high sparsity in attention maps inevitably aggravates the extent of both irregular data accesses and processing, which could incur severe workload imbalance problems.
\hry{Moreover, the high sparsity can cause undesired under-utilization when processing highly sparse attention regions, where efficiency is largely bounded by memory/bandwidth due to decreased computational density and severe off-chip traffics, as indicated by the roofline model in Fig. \ref{fig:roofline}. 
That is because the non-zero elements in sparse attention maps of ViTs mostly concentrate along the diagonal lines, as shown in Fig. \ref{fig:split_and_conquer}, which is actually the most inefficient pattern since it requires loading all corresponding $Q$ (query vectors in attentions) and $K$ (key vectors in attentions) vectors into the on-chip memory for calculating only a small amount of attention scores ($S=Q \cdot K^T$). Hence, 
such diagonal sparse patterns cause not only reduced reuse opportunities of loaded $Q/K$ vectors but also a low PE utilization dilemma, i.e., a high sparsity ratio can help reduce the computation, but also causes a large data movement bottleneck, which can greatly compromise the achievable efficiency of accelerating ViTs.
}

To better leverage the new opportunities of ViTs' fixed sparse patterns and overcome their challenges of workload imbalance and low utilization, this work targets a dedicated ViT acceleration solution for maximizing the achievable efficiency. Specifically, we make the following contributions:

\begin{itemize}
\setlength\itemsep{0.1em}
  \vspace{-0.2em}
    \item We propose a \textbf{ViT} algorithm-accelerator \textbf{Co}-\textbf{D}esign framework dubbed ViTCoD, aiming to leverage ViTs’ unique opportunities and to tackle ViTs specific acceleration bottlenecks to boost ViTs' acceleration efficiency by harmonizing algorithm- and accelerator-level innovations. 
    To the best of our knowledge, ViTCoD is the first co-design framework dedicated to accelerating \hr{sparse} ViTs' inference, offering a new perspective on efficient ViT solutions.

\begin{figure}[t]
    \centering
    \includegraphics[width=0.9\linewidth]{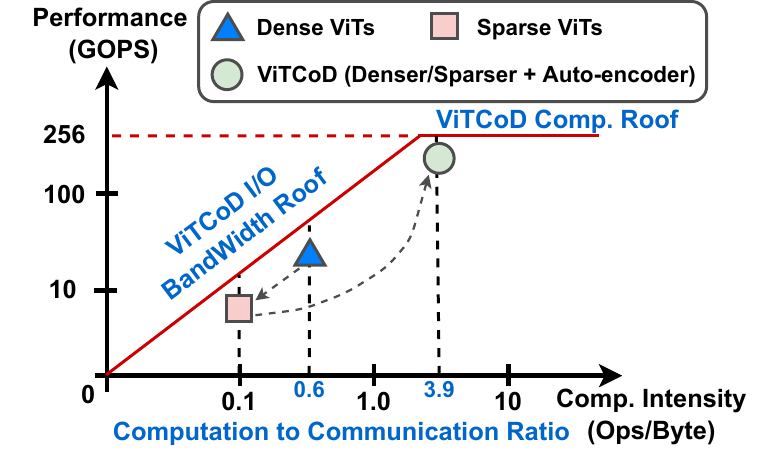}
    \vspace{-0.6em}
    \caption{\hry{Roofline model analysis for ViTCoD when only accelerating the key attention bottlenecks ($S = Q \cdot K^T$).
    Here \textbf{Dense ViTs} refer to the original dense attention workload, while \textbf{Sparse ViTs} refer to polarized denser/sparser attention worloads. Both of them require to load all $Q/K$ vectors and are limited by the bandwidth (even worse for sparse ViTs due to the reduced computation). The \textbf{ViTCoD} adopts not only sparse attention but also an auto-encoder module to largely reduce the communication towards better designs.
    }}
    \vspace{-2em}
    \label{fig:roofline}
\end{figure}

    \item On the algorithm level, 
    \hry{
    ViTCoD prunes and polarizes the attention maps to make them have either denser or sparser fixed patterns for regularizing two levels of workloads without severely hurting the accuracy, largely reducing the dominant attention computations.
    }
    On top of that, we further integrate a lightweight and learnable auto-encoder module to enable trading the dominant high-cost data movements for lower-cost computations. 
    \hry{
    Specifically, to alleviate the low PE utilization problem, ViTCoD's auto-encoder module helps to reduce the computation-to-communication ratio, leveraging
    a hypothesis that $Q$ and $K$ vectors among different attention heads have similarities and thus can be recovered from a much compressed representation.
    Hence, ViTCoD algorithm enables highly sparse attentions while enhancing regular and reduced data accesses, pushing memory/bandwidth bounded scenarios towards optimal designs (see Fig. \ref{fig:roofline}'s roofline analysis).
    }

    \item On the hardware level, 
    we develop a dedicated two-pronged accelerator that can simultaneously coordinate the aforementioned enforced denser and sparser workloads for boosted hardware utilization, while integrating on-chip encoder and decoder engines to leverage ViTCoD's algorithm pipeline for much reduced data movements. Specifically, one branch accelerates the polarized denser patterns with enhanced regular data accesses, while the other branch accelerates (mostly on-chip) the remaining irregular but largely reduced sparser workloads; The encoder and decoder engines are leveraged to compress the Q/K vectors before transferring them back to off-chip memory and then recover them when being loaded into the on-chip memory, trading high-cost data movements for low-cost computations to boost efficiency.

\begin{table*}[t]
    \centering
    \caption{\hry{A taxonomy for classifying and comparing representative sparse accelerators.}}
    \vspace{-0.7em}
    \resizebox{\linewidth}{!}{
  \begin{tabular}{l||c|c|c|c|c|c|c}
  \hline 
    & \textbf{OuterSpace~\cite{pal2018outerspace}} & \textbf{ExTensor~\cite{hegde2019extensor}} & \textbf{SpArch~\cite{zhang2020sparch}} & \textbf{Gamma~\cite{zhang2021gamma}} & \textbf{SpAtten~\cite{wang2021spatten}} & \textbf{Sanger~\cite{lu2021sanger}} & \textbf{ViTCoD (\textcolor{blue}{Ours})} \\
    \hline
  \hline
  \textbf{Application Field} & Tensor Algebra & Tensor Algebra & Tensor Algebra & Tensor Algebra & NLP Transformer & NLP Transformer & ViT  \\
  \hline
  \textbf{Workloads} & SpGEMM & SpGEMM & SpGEMM & SpGEMM &
  \tabincell{c}{Sparse Attention:\\SDDMM; SpMM} &
  \tabincell{c}{Sparse Attention:\\SDDMM; SpMM} &
  \tabincell{c}{Sparse Attention:\\SDDMM; SpMM} \\
  \hline
  \textbf{Dataflow} & 
  \tabincell{c}{Outer-product\\(Input-stationary)} &
  \tabincell{c}{Hybrid Outer-product\\\& Inner-product (Input-\\\& Output-stationary)} &
  \tabincell{c}{Condensed\\ Outer-product\\(Input-stationary)} &
  \tabincell{c}{Gustavson(Row)-\\stationary} &
  Top-k Selection & S-stationary &
  \tabincell{c}{K-stationary;\\Output-stationary} \\
  \hline
  \tabincell{l}{\textbf{Sparsity Pattern}} &
  Static & Static & Static & Static &
  \tabincell{c}{Dynamic \& \\Input-dependent} &
  \tabincell{c}{Dynamic \& \\Input-dependent} & Static \\
  \hline
  \tabincell{l}{\textbf{Pattern Regularity}} &
  Unstructured & Unstructured & Unstructured & Unstructured &
  \tabincell{c}{Coarse-grained \& \\Structured} &
  \tabincell{c}{Fine-grained \& \\Structured} & Denser \& Sparser \\
  \hline
  \textbf{Off-chip Traffic} & High & Low $\sim$ Medium & Low $\sim$ Medium & Low & Medium & High & Low  \\
  \hline
  \textbf{Bandwidth Requirement} & Medium & Medium $\sim$ High & Low & Low & Medium $\sim$ High & Medium $\sim$ High & Low  \\
  \hline
  \textbf{Sparsity} & High $\sim$ Ultra High & High $\sim$ Ultra High & High $\sim$ Ultra High & High $\sim$ Ultra High & Low & Medium & High  \\
  \hline
  \textbf{Alg. \& HW Co-design} & \cmark & \xmark & \xmark & \xmark & \cmark & \cmark & \cmark  \\
  \hline
  \end{tabular}
    }
    \label{tab:taxonomy}
    \vspace{-1.3em}
\end{table*}

    \item Extensive experiments and ablation studies on various ViT models consistently validate the advantages of our proposed ViTCoD framework, leading to
    \hr{235.3$\times$, 142.9$\times$, 86.0$\times$, 10.1$\times$, and 6.8$\times$} speedups over both general computing platforms CPUs, EdgeGPUs, GPUs, and prior-art Transformer accelerators SpAtten and Sanger, respectively, while maintaining the model accuracy.

\end{itemize}

%% file: sections/2-Related_Work.tex
\section{Related Works}
\label{sec:related_works}
\vspace{-0.1em}
\textbf{Vision Transformers (ViTs).}
Motivated by Transformers' strong representation capabilities for NLP tasks~\cite{vaswani2017attention}, there has been a growing interest in developing Transformers for CV tasks.
Specifically, inspired by the self-attention mechanism, \cite{hu2018squeeze,zhang2020resnest} propose novel attention mechanisms for CNNs; \cite{bello2019attention,wu2020visual} integrate Transformer and CNN within the same model; \cite{vit,pmlr-v119-chen20s} design pure Transformer architectures for CV tasks.
Among these exploration, Vision Transformer (ViT)~\cite{vit} adopts a simple and intuitive architecture design by splitting input images into small patches and directly applying pure Transformers to those patches~\cite{deng2009imagenet}.
Later, DeiT~\cite{touvron2020training} proposes an improved ViT training recipe 
~\cite{deng2009imagenet}, and achieves a comparable accuracy without the necessity of costly pretraining.
To further improve the accuracy or efficiency on vision tasks,
CrossViT~\cite{chen2021crossvit},
PiT~\cite{heo2021rethinking}, PVT~\cite{wang2021pyramid}, and Swin-Transformer~\cite{liu2021Swin} propose a pyramid-like architecture for designing ViTs, which is commonly used in CNNs~\cite{Dai2020FBNetV3JA,howard2019searching,xiong2020mobiledets}.
With the goal of deploying ViT in resource-constrained devices, prior works have also explored more efficient ViTs from different perspectives.
For example, 
LeViT~\cite{graham2021levit}, CvT\cite{wu2021cvt}, and MobileViT~\cite{mehta2021mobilevit} propose more efficient self-attention implementation or incorporate convolutional feature projection blocks into ViTs.
Different from those works, our ViTCoD
contributes a new systematic way to explore the possibility of both fixed ViTs sparse attentions and balanced data movements and computations from both algorithm and hardware levels, without largely degrading the model accuracy.

\textbf{Sparse Attention Algorithms.} 
As commonly recognized, the computational complexity of self-attention in Transformers is quadratic to the 
length of the sequences (or the total number of patches in the input images for ViTs)~\cite{kitaev2020reformer,zhu2021long}.
To make the attention module more efficient, 
there have been a number of attempts to build sparse attention algorithms.
For example,
for NLP Transformers,
BigBird~\cite{zaheer2020big} constructs attention maps by merging random-, window-, and global-attentions together while keeping the remaining parts to be zeros;
Reformer~\cite{kitaev2020reformer} uses locality sensitive hashing to compute the nearest neighbors instead of all tokens in the attentions;
BlockBERT~\cite{qiu2019blockwise} proposes the block sparsity for the attention map to reduce the complexity;
BigBird~\cite{zaheer2020big} and BlockBERT~\cite{qiu2019blockwise} propose the structured or block sparsity for the attention map while requiring to predict dynamic and input-dependent sparse patterns.
While NLP Transformers require to predict dynamic and input-dependent sparse attention patterns to achieve a medium sparsity around 50\%.
For ViTs,
\cite{Kim_2021_CVPR} is one of the first works to explore sparse attentions and provides a thorough visualization and analysis of attention patterns, showing the feasibility of adopting fixed sparse masks for ViTs while achieving a high sparsity (e.g., 90\%). 
In contrast, ViTCoD is the first algorithm and accelerator co-design framework dedicated to accelerate sparse ViTs, that fully exploits the fixed sparse patterns from both algorithm and hardware perspectives.

\textbf{\hry{Sparse Tensor Algebra Accelerators.}}
\shh{Sparse General Matrix Multiplication (SpGEMM)
, commonly used in machine learning algorithms, 
is known to be hardware unfriendly to general-purpose platforms (e.g., CPUs and GPUs), and thus calls for dedicated accelerators \cite{pal2018outerspace, hegde2019extensor, zhang2020sparch, zhang2021gamma,Srivastava2020MatRaptorAS,Srivastava2020TensaurusAV} to explore different dataflows for alleviating poor data locality and dedicated accelerators, as summarized in Table \ref{tab:taxonomy}.
Furthermore, MatRaptor \cite{Srivastava2020MatRaptorAS} leverages a row-wise dataflow to promote data reuses, and a novel sparse storage format to boost memory bandwidth utilization; 
Additionally, Tensaurus \cite{Srivastava2020TensaurusAV} proposes a new sparse storage format for sparse tensor kernels, and develops an accelerator to simultaneously support sparse/dense tensor factorizations as well as common mixed sparse-dense matrix operations.
Overall, existing sparse tensor algebra accelerators are mainly dedicated for SpGEMM workloads with unstructured patterns. In contrast,  
ViTCoD algorithm integrates a learnable auto-encoder module to efficiently alleviate the memory/bandwidth bottleneck of ViTs attentions' sample-based dense-dense matrix mulitplication (SDDMM), and our ViTCoD framework contributes the first dedicated co-design for accelerating ViTs' sparse attention (denser/sparser) workloads, of which the overall design space exploration can provide insights for developing efficient ViT solutions.}

\textbf{Existing Transformer Accelerators.}
The unique execution patterns of the costly self-attention modules have motivated dedicated accelerator designs or algorithm and accelerator co-designs for Transformers.
For example,
A$^3$~\cite{ham2020a3} is the first work to approximate the attention by greedily searching for $K$ vectors that are relevant to the current $Q$ vector to reduce the amount of computations, of which the approximation hurts the model accuracy at high sparsity levels;
ELSA~\cite{ham2021elsa} 
approximate the attention
by directly using binary hashing maps to estimate the angle between $Q$ and $K$ vectors at a cost of non-negligible accuracy drops;
SpAtten~\cite{wang2021spatten} structurally removes unnecessary attention heads and input tokens, which is therefore coarse-grained and leads to a low achievable sparsity ratio;
Sanger~\cite{lu2021sanger} adopts low precision $Q$ and $K$ vectors for estimating the sparse attention masks, which are then packed and split to be more regular and friendly supported by a reconfigurable architecture.
DOTA~\cite{qu2022dota} considers both low precision and low rank linear transformation to predict the sparse attention masks, and explores token-level parallelism and out-of-order execution for locality-aware computing.
All above works focus on NLP Transformers, and thus require dynamic and input-dependent sparse masks prediction.
All accelerators above targeting NLP Transformers require dynamic and input-dependent sparse masks prediction.
For accelerating ViTs, 
VAQF~\cite{sun2022vaqf} designs inference accelerators on FPGAs for quantized ViTs with binary weights and low precision activations; 
In contrast, ViTCoD is the first algorithm and accelerator co-design framework dedicated to sparse ViTs, aiming to fully exploit ViTs' fixed sparse patterns and incorporating an auto-encoder module to boost sparse ViTs' utilization.

%% file: sections/3-Methods.tex
\begin{figure}[t]
    \centering
    \includegraphics[width=\linewidth]{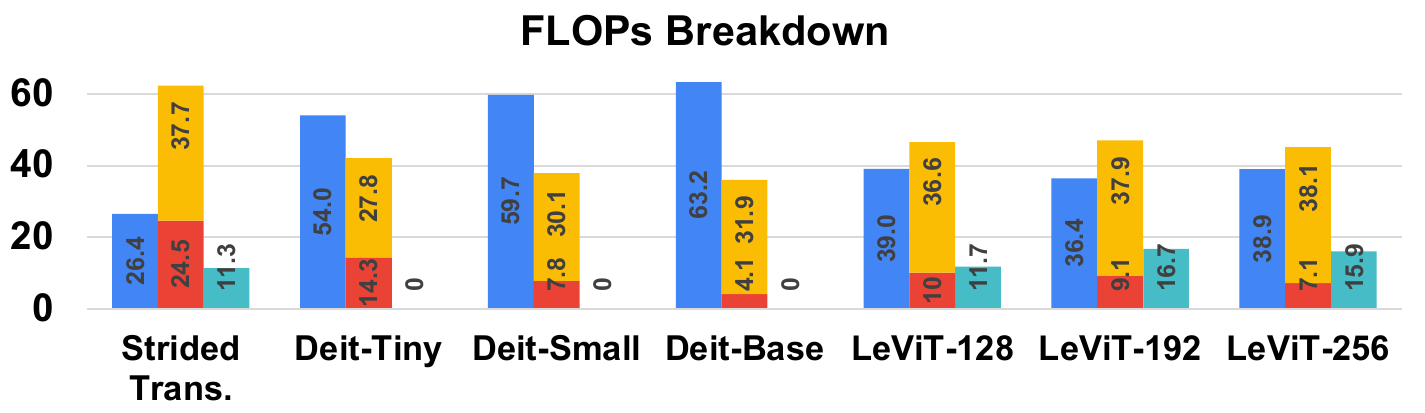}
    \includegraphics[width=\linewidth]{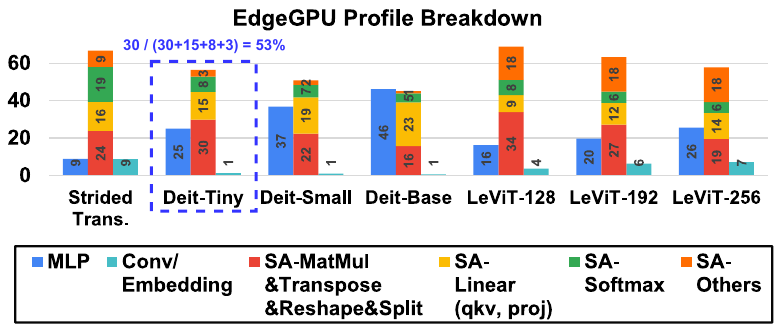}
    \vspace{-2em}
    \caption{\hrhpca{The FLOPs (top) and measured latency (bottom) breakdowns of various ViTs on an EdgeGPU TX2 \cite{edgegpu}, where the self-attention (SA) module denoted by middle bars accounts for over 50\% of the total latency.}}
    \label{fig:edgegpu}
    \vspace{-1.5em}
\end{figure}

\begin{figure*}[t]
    \centering
    \includegraphics[width=\linewidth]{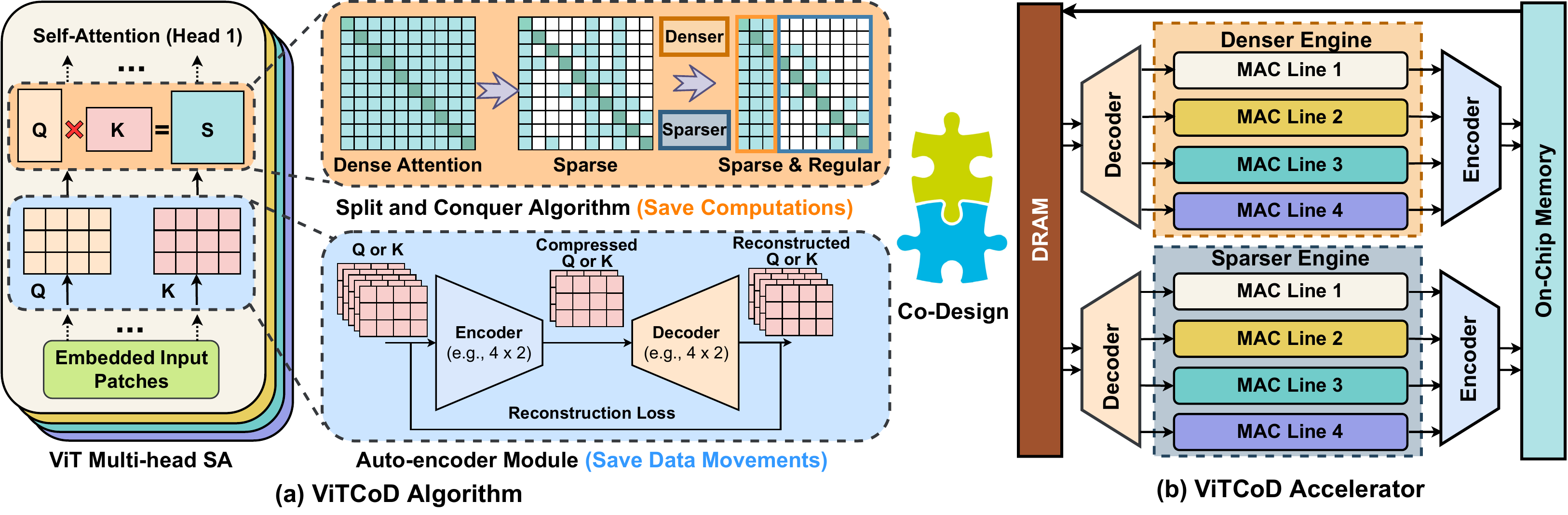}
    \vspace{-1.8em}
    \caption{An overview of ViTCoD, the first algorithm-accelerator co-design framework dedicated to sparse ViTs.}
    \vspace{-1.2em}
    \label{fig:overview}
\end{figure*}

\section{ViTCoD: Motivation \& Overview}

\subsection{Bottlenecks in ViT Inference}
\label{sec:bottlenecks}

To better understand the bottleneck in ViT inference, we measure and summarize the FLOPs and end-to-end latency breakdown for ViT inference measured on commercial edge devices, as shown in Fig. \ref{fig:edgegpu}. In this set of profile experiments, we consider various ViT models, including both (1) standard DeiT~\cite{DeiT}, LeViT~\cite{graham2021levit} for mobile scenarios and (2) Strided Transformer~\cite{li2022exploiting} achieving SOTA performance on AR/VR applications. \hrhpca{From Fig. \ref{fig:edgegpu}, we can see that although the self-attention module is not as dominant as MLPs in terms of FLOPs,}
it consistently accounts for over 50\% of the total latency (as high as 69\% in LeViT-128~\cite{graham2021levit}) when being executed on real mobile devices.
Moreover, the matrix multiplications among $Q/K/V$ vectors (i.e., $Q \cdot K^T$ and $S \cdot V$) and their corresponding reshape/split operations occupy up to 53\% latency of the self-attention module on EdgeGPU platforms \cite{edgegpu}.
Note that this set of breakdown statistics is consistent with that of NLP Transformers as mentioned in \cite{wang2021spatten}, indicating that the self-attention module, especially its core matrix multiplications among $Q/K/V$ vectors, is indeed a major bottleneck in Transformer inference acceleration.
The above bottleneck analysis indicates that there exists a fundamental dilemma associated with ViT inference acceleration:
\hry{
On one hand, to boost ViTs' inference efficiency, it is desired that the attention maps are (if not highly) sparse, which 
might lead to more irregular data accesses of the $Q/K/V$ vectors.}
\hry{Moreover, for highly sparse attentions, it is likely that the acceleration largely bottlenecked by data movements, causing low utilization of PEs, as also analyzed in Fig. \ref{fig:roofline}.}
On the other hand, maintaining ViTs' dense attention maps and thus task accuracy is likely to require a higher hardware cost for ViT inference as discussed in recent studies \cite{lu2021sanger,ham2021elsa,wang2021spatten}, limiting their more extensive applications.

\subsection{ViTCoD Overview}

Fig. \ref{fig:overview} illustrates an overview of the proposed ViTCoD, which aims to alleviate the aforementioned dilemma and thus the self-attention bottleneck.
Specifically, we leverage (1) a split and conquer algorithm to prune the attention maps by up to 90\% sparsity and to simultaneously polarize the attention maps to be either denser or sparser for enhancing more regular workloads; and (2) an auto-encoder module to compress the corresponding vectors for calculating attentions to a much more compact representation, without hurting the model accuracy.
On top of that, we further design a dedicated two-pronged accelerator integrating the encoder and decoder engines to (1) cooperatively handle the denser or sparse workloads and (2) leverage the auto-encoder module of ViTCoD's algorithm pipeline for much reduced data movements.
%
Next, we will introduce the ViTCoD algorithm and accelerator in detail.


\section{Proposed ViTCoD Algorithm}

\subsection{Preliminaries of Self-Attention and ViTs}
\label{sec:alg_preliminary}

\begin{figure}
    \centering
    \includegraphics[width=0.95\linewidth]{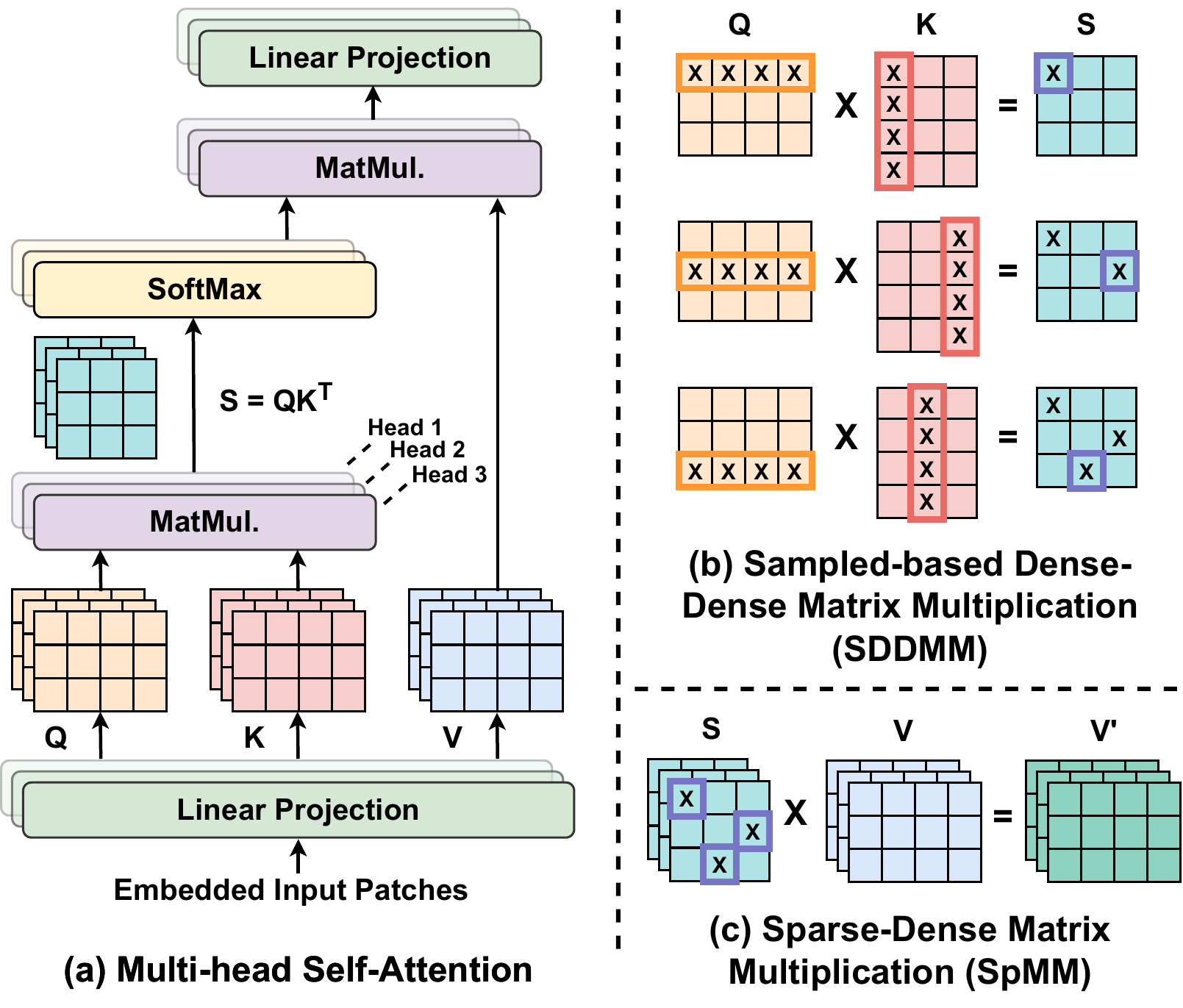}
    \vspace{-1.2em}
    \caption{Illustrating the self-attention workflow and its associated matrix multiplication patterns.}
    \vspace{-1em}
    \label{fig:self_attention}
\end{figure}

\textbf{Self-Attention.}
Self-attention is a core component of Transformers \cite{transformer}, and consists of a number of heads $\text{H}$ with each capturing different global-context information via measuring pairwise correlations among all tokens as illustrated in Fig. \ref{fig:self_attention} (a) and defined below:
    \begin{equation} \label{equ:attn_op}
    \begin{split}
    \textit{\textbf{O}}_{\tt \textbf{Attn}} \!=\! {\tt Concat}(\text{H}_1, \cdots, \text{H}_h) \cdot W^O,
    \,  \mathrm{where} \,\, \\ 
    \text{H}_i \!=\! {\tt Softmax}(\frac{QW_i^Q \cdot (KW_i^K)^T}{\sqrt{d_k}}) \cdot VW_i^V,
    \end{split}
\end{equation}
where $h$ denotes the number of heads, $Q, K, V \in \mathbb{R}^{n \times d}$ are the query, key, and value vectors of hidden dimension $d$ obtained by linearly projecting the input sequence of length $n$, respectively.
For each head, $W_i^Q, W_i^K, W_i^V \in \mathbb{R}^{d \times d_k}$ are learnable projection weight matrices, where $d_k = d / h$ is the embedding dimension of each head. In this way, the attention block first computes dot-products between the key-query pairs, then scales the dot-product results to stabilize the training, uses ${\tt Softmax}$ to normalize the resulting attention scores, and finally computes a weighted sum of the value embeddings corresponding to different inputs.
Finally, the results from all the heads are concatenated and further projected with a weight matrix $W^O \in \mathbb{R}^{d \times d}$ to generate the final outputs.

To alleviate the bottleneck computational complexity from the above self-attention mechanism, sparse attention techniques have emerged as a promising solution, under which the introduced sparsity results in the following two kinds of the attention matrix multiplication:
(1) the first multiplication between $Q$ and $K$ becomes a general sampled dense-dense matrix multiplication (SDDMM) based on the location of the nonzero samples in the attention maps, as illustrated in Fig. \ref{fig:self_attention} (b);
and (2) the following multiplication between the attention maps and the V matrix becomes a sparse-dense matrix multiplication (SpMM), as illustrated in Fig. \ref{fig:self_attention} (c).
As such, they inevitably lead to irregular workload patterns when accessing the Q, K, and V matrices, making it challenging to parallelize the computation and causing temporal load imbalance if the non-zero attention scores are not evenly distributed.

\begin{figure}[t]
    \centering
    \includegraphics[width=0.95\linewidth]{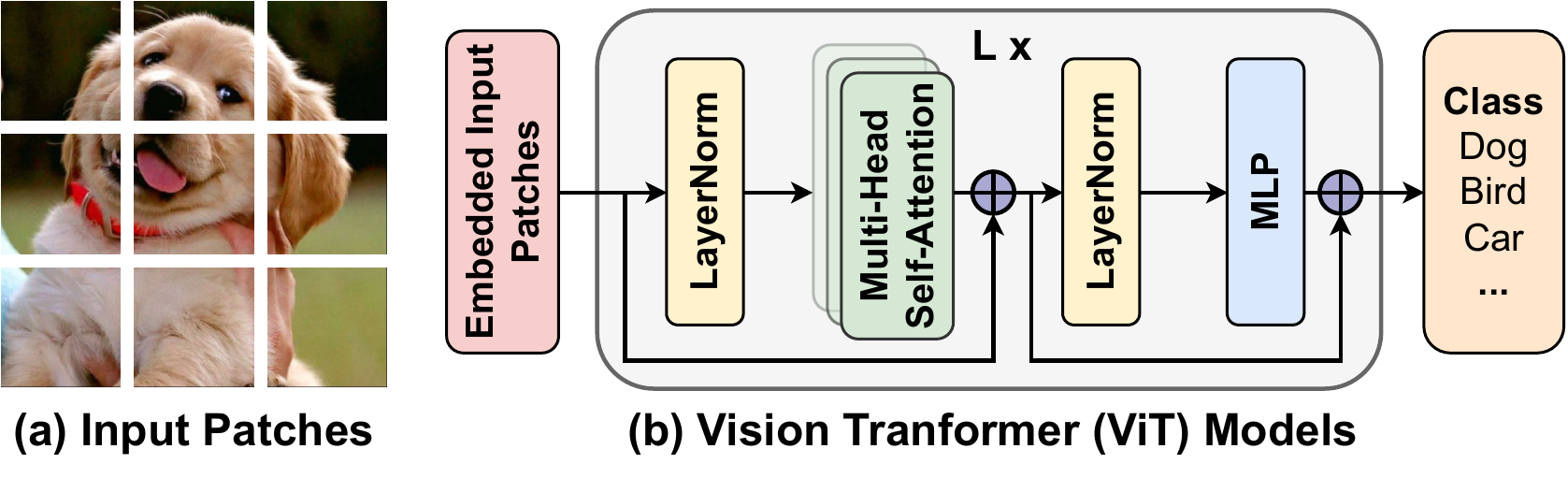}
     \vspace{-1.2em}
    \caption{Illustrating the input patches and ViT models.}
     \vspace{-1.5em}
    \label{fig:ViT}
\end{figure}

\textbf{ViT Models and Variants.}
As illustrated in Fig. \ref{fig:ViT} (b),
ViT models (e.g., DeiT \cite{DeiT}) consist of alternating layers of multi-head self-attention (MHSA) and multi-layer perceptron (MLP) blocks, where MLP contains two fully-connected layers with a non-linearity function of Gaussian error linear unit (GELU). Additionally, LayerNorm (LN) is applied before every block.
During training or inference, the input images are split into patches of fixed size as shown in Fig. \ref{fig:ViT} (a), and then each patch will be linearly projected into embedded patches before being fed into the ViT models as a sequence of vectors.
Recently, a surge of research works target to build ViT variants for resource-constrained devices, 
e.g., LeViT \cite{graham2021levit} achieves much higher efficiency by using a multi-stage Transformer architecture and a few convolution layers that are incorporated before the ViT blocks. 
For this paper, we mainly focus on the ViT blocks as the early convolutions only account for a negligible amount of FLOPs (i.e., $<$ 7\%).

\subsection{ViTCoD's Split and Conquer Algorithm}
\label{sec:alg_split_and_conquer}


\textbf{Design Considerations.}
Our ViTCoD's split and conquer algorithm aims to alleviate the costly (e.g., over 50\% of total latency as shown in Fig. \ref{fig:edgegpu}) quadratic computational complexity to the number of input tokens/patches in self-attention blocks, by enforcing fixed sparse attention masks with merely two levels of computation workloads for the ease of acceleration. Specifically, we adopt both pruning with fixed masks and attention map reordering:

\vspace{-0.2em}
\begin{itemize}
\setlength\itemsep{0.1em}
    \item \textit{Pruning with Fixed Masks:} 
    As mentioned, ViTs enjoy a relatively fixed number of input tokens/patches for all inference images,
    while NLP Transformers have to handle inputs with various sequence lengths. Such a discrepancy makes previous accelerators not optimal for accelerating ViTs, and leaves room for designing fixed sparse attention pattern of high sparsity without hurting ViTs' accuracy.
    To generate the desired sparse mask patterns, we first extract averaged attention maps by forwarding the pretrained models on all training samples,
    and then perform pruning according to a criterion of the remaining information quantity.
    In particular, for each query, we select only attentions of high value by pruning the remaining when the cumulative sum of the sorted and normalized attention scores of descending order is equal or greater than $\theta_p$, where $\theta_p$ is the predefined threshold for evaluating the information quantity. 
    Such pruning will generate a binary mask for each attention map, where ``1” and ``0” denote the reserved and pruned attentions, respectively.

    \item \textit{Attention Map Reordering:}
    To reduce the resulting sparse masks' irregularity from the above pruning, we identify and cluster the query-key (Q/K) pairs into two patterns as shown in Fig. \ref{fig:split_and_conquer}, resulting in merely two levels (i.e., either denser or sparser) of computation workloads for favoring the acceleration efficiency.
    Specifically, the denser pattern embodies global tokens that have high correlations with all other tokens.
    On the other hand, the sparser pattern refers to the remaining sparse masks except for the diagonal lines, where most locations are of ``0” values. This intriguing phenomenon is formed because adjacent input tokens/patches tend to have a higher correlation than others.

\end{itemize}


{
\begin{figure}[!t]

\makeatletter
\newcommand{\removelatexerror}{\let\@latex@error\@gobble}
\makeatother

\begin{minipage}{0.477\textwidth}
\centering

\begingroup
\removelatexerror

\begin{algorithm}[H]

\caption{ViTCoD's \!\! Split \!\! \& \!\! Conquer \!\! Algorithm.}
\label{alg:split_and_conquer}

\SetAlgoLined
\KwIn{Normalized attention map $A\in\mathbb{R}^{n\times n}$, pruning threshold $\theta_p$, dense threshold $\theta_d$} 
\KwOut{Pruned \& reordered attention map $m \odot A'$, the number of global tokens $N_{gt}$}


\CommentL{\textcolor{purple}{\textit{// Pruning with Fixed Masks}}}

\tikzmkc{A}
$Sum = 0; \,\, idx_p = 0; \,\, order = Argsort(A);$ 

\While(\Comment*[f]{Find Masks}){$Sum < \theta_p$ and $idx_p < n^2$}{
    
    $Sum \leftarrow Sum + Sort(A)[idx_p];$ 
    
    $idx_p \leftarrow idx_p + 1;$

}

$m = order[Argsort(order) <= idx_p].$ \Comment*[f]{Masks}
\tikzmkc{B} \boxitc{orange}

\CommentL{\textcolor{blue}{\textit{// Attention Map Reordering}}}

\tikzmka{A}
$N_{gt} = 0; \,\, idx_d\leftarrow[1,2,\cdots,n];$

\For(\Comment*[f]{Find Global Tokens}){$i = 0$ to $n-1$}{
    \If{$\|(m \odot A)_{:,i}\|_0 > \theta_d$}{
    
        $\textsc{Swap}(idx_d[N_{gt}], \,\, idx_d[i]);$
        
        $N_{gt} \leftarrow N_{gt} + 1;$
    }
}

$A'\leftarrow\textsc{Permute}(A, \,\, idx_d).$ \Comment*[f]{Reorder A}
\tikzmkc{B} \boxita{cyan}

\Return $m \odot A'$ and $N_{gt}.$ 

\end{algorithm}

\endgroup

\end{minipage}
\vspace{-1.5em}
\end{figure}
}

\textbf{The Split and Conquer Algorithm.}
We describe the split and conquer algorithm in Alg. \ref{alg:split_and_conquer}.
For a given averaged and normalized attention map $A\in\mathbb{R}^{n\times n}$ extracted from a pretrained ViT model on all training samples, we prune and reorder it into either the denser or sparser patterns.
Specifically, we first prune the attention maps based on a predefined threshold $\theta_p$ to generate the fixed masks \hry{\textbf{(Line 1-6)}}, where the decreasingly sorted attention scores will be accumulated until their sum reaches $\theta_p$ \hry{\textbf{(Line 3)}} and then the indexes associated with those accumulated attention scores will be kept and set to ``1'' in the corresponding binary mask, while leaving the remaining as ``0'' \hry{\textbf{(Line 6)}}.
The second step is to reorder the sparse attention map for enforcing the desired patterns \hry{\textbf{(Line 7-14)}}. Specifically, we move those tokens with a larger value of non-zero elements than a predefined threshold $\theta_d$ (denoted as global tokens) to the front as the denser pattern \hry{\textbf{(Line 10)}}, while treating the remaining ones as the sparser pattern. 
\hry{Note that here the operator with a circle and dot means element-wise multiplication; \textsc{SWAP} means exchanging the index so that all global tokens can be moved to the leftmost w.r.t such indexes by executing \textsc{PERMUTE}.}
As a result, both the resulting reordered attention map $A'$ \hry{\textbf{(Line 14)}} and the number of global tokens $N_{gt}$ favor the ease of hardware acceleration.
Finally, we will finetune the resulting ViT model with pruned and reordered attentions $m \odot A'$ to restore the model accuracy. Note that the sparse attention masks will remain \textit{fixed} during both finetuning and inference for favoring efficient inference.

\begin{figure}[t]
    \centering
    \includegraphics[width=\linewidth]{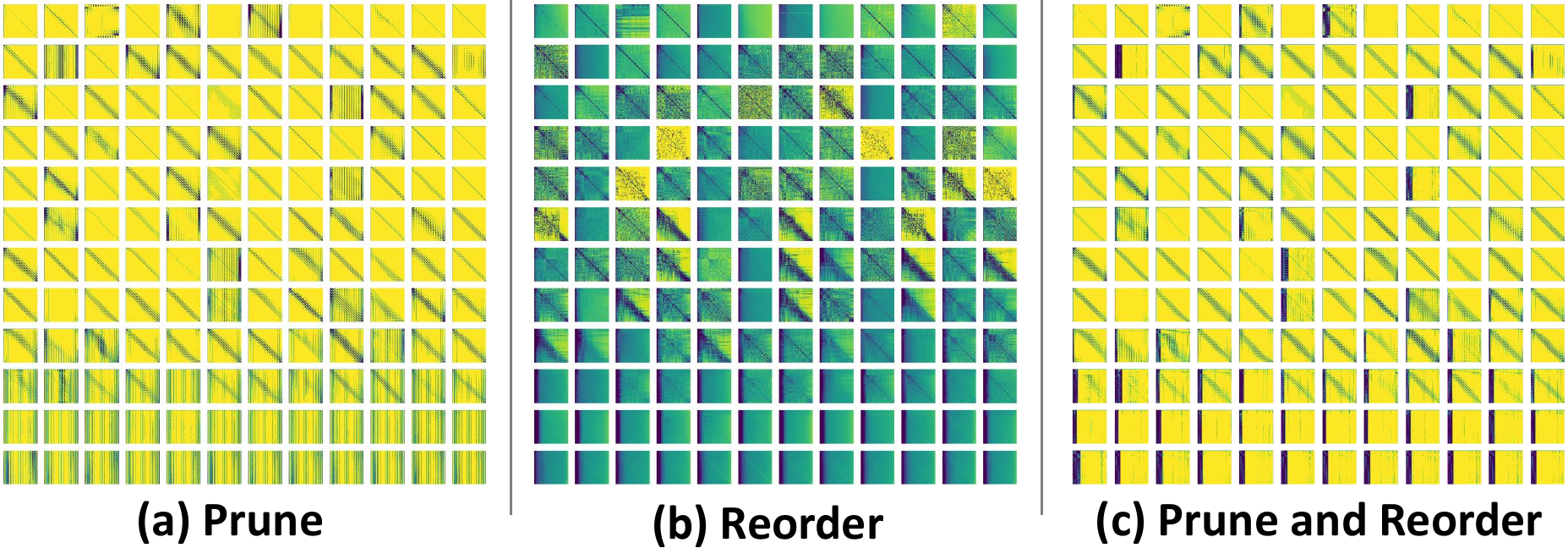}
    \vspace{-1.8em}
   \caption{\hrhpca{Visualizing the attention maps of 144 heads (12 layers$\times$12 heads) in DeiT-Base after gradually applying our split and conquer algorithm: (a) pruning only; (b) reordering only; (3) both pruning and reordering, where the resolution is 197$\times$197 with 197 being the number of input tokens/patches.}}
    \vspace{-1.5em}
    \label{fig:attn_visual}
\end{figure}


\textbf{Visualization of Attention Maps.}
To validate the effectiveness of the proposed split and conquer algorithm, we visualize the resulting attention maps of all 12 layers (12 heads in each layer) from DeiT-Base \cite{DeiT} in Fig.~\ref{fig:attn_visual}. 
We can clearly observe the improved regularity. In particular, there exists a notably clustered dense block at the left side of most attention maps, which can be accelerated by a dense computing engine with high utilization; the remaining attentions are very sparse, located either on the diagonal lines or uniformly distributed among the whole attention map.
Such patterns create new opportunities for dedicated accelerators to fulfill the promise of the extreme sparsity into real-hardware efficiency.

\vspace{-0.3em}
\subsection{ViTCoD Learnable Auto-encoder Module}
\label{sec:alg_auto_encoder}

\textbf{Design Considerations.}
Although we have enforced the two desired workloads from the above split and conquer algorithm, the sparser workload can still suffer from a low PE utilization dilemma. That is, a higher sparsity helps to reduce the amount of attention computations, but a large data movement bottleneck still exists due to the diagonal concentration of the non-zero attention values. As analyzed in the roofline model (See Fig. \ref{fig:roofline}),
such a diagonal pattern is actually the most inefficient, 
To tackle this issue,
if we naively shrink the dimension of $Q$ and $K$, then the resulting attention maps can suffer from the associated low-rank approximation, i.e., $rank(S) < min(rank(Q), rank(K))$, degrading the achievable accuracy, as widely discussed in \cite{bhojanapalli2020low}.

\begin{figure}[t]
    \centering
    \includegraphics[width=\linewidth]{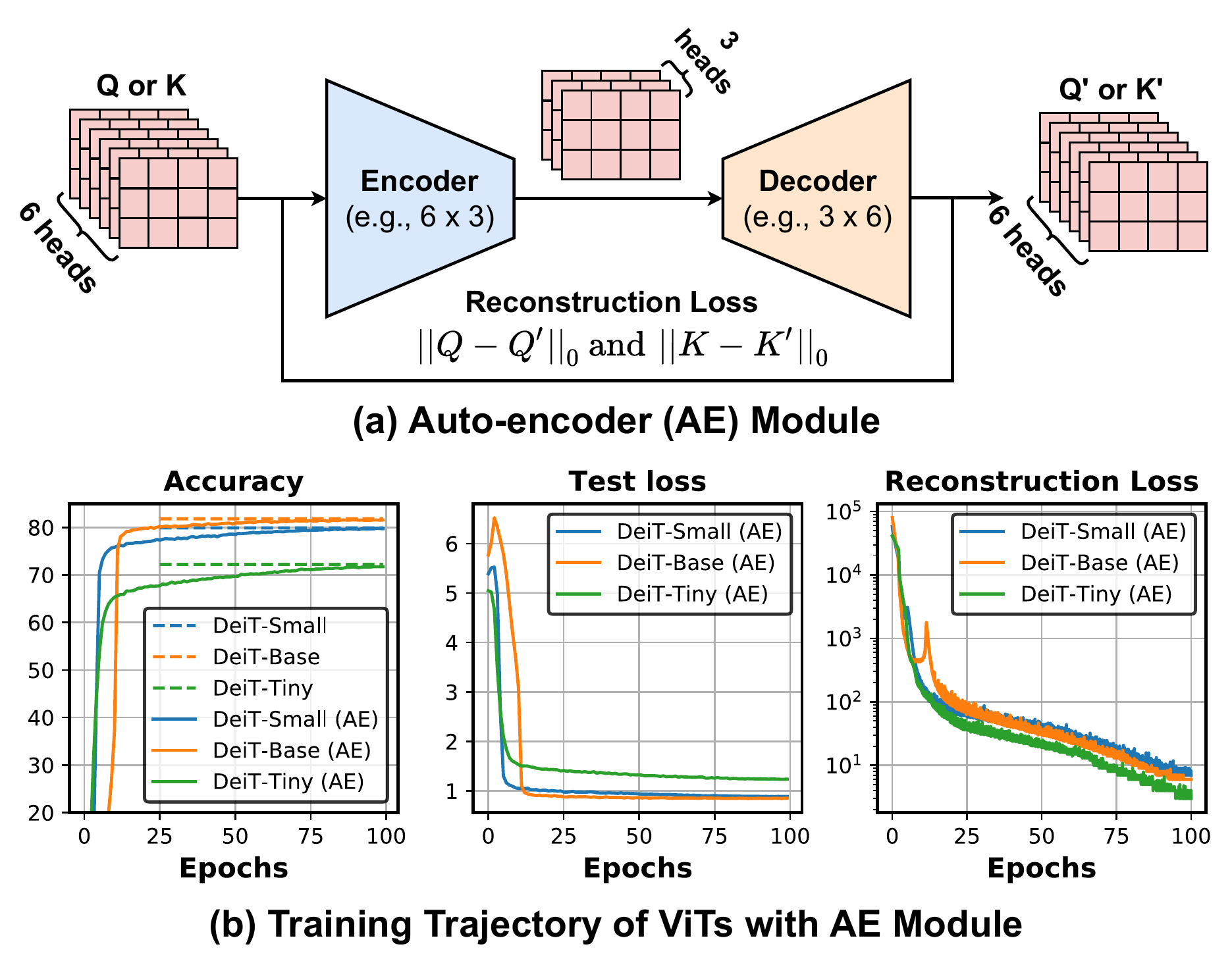}
    \vspace{-2em}
   \caption{Illustrating (a) the lightweight and learnable auto-encoder module and (b) the training trajectory of ViTs with such auto-encoder modules, where the dashed lines denote the corresponding vanilla/original ViTs' accuracy without any compression.}
    \label{fig:auto_encoder}
    \vspace{-1.5em}
\end{figure}

To overcome the above dilemma, we propose to design a lightweight and learnable auto-encoder module that compresses the Q/K vectors to a much more compact representation before they are moved to the off-chip memory, and then recover them back after they are loaded into the off-chip memory. 
The hypothesis is that although the Q/K dimension cannot be reduced, there is still a large degree of redundancy among different heads.
As illustrated in Fig. \ref{fig:auto_encoder} (a), we 
use a lightweight encoder to compress the Q/K vectors along the attention head dimension, and then adopt another decoder to recover them back as Q'/K'.
To enforce the recovered Q'/K' to be as close to the original Q/K as possible, we leverage their discrepancy (e.g., $|| Q - Q' ||_0$) as a reconstruction loss, so that the auto-encoder module is learnable and will be jointly optimized together with the ViT model weights. In this way, the data movements associated with accessing the $Q$ and $K$ vectors from the costly off-chip memory are largely alleviated, while the model accuracy is maintained.

\textbf{ViT Training with Auto-encoder Modules.}
After inserting the aforementioned auto-encoder modules to a pretrained ViT model, we will finetune the resulting model for jointly training the auto-encoder and model weights. The overall objective loss function is as follows:
\vspace{-0.6em}
\begin{equation}\label{eq:joint_train}
    \mathcal{L} = \mathcal{L}_{CE} + \mathcal{L}_{Recons} = \mathcal{L}_{CE} + || Q - Q' ||_0 + || K - K' ||_0
\end{equation}
\vspace{-0.2em}
where $\mathcal{L}_{CE}$ denotes the cross-entropy test loss while $\mathcal{L}_{Recons}$ denotes the reconstruction loss.
We visualize the training trajectory in Fig. \ref{fig:auto_encoder} (b), from which we can see that
(1) both the test loss (i.e, $\mathcal{L}_{CE}$) and reconstruction loss are significantly reduced, verifying the convergence and effectiveness of our proposed auto-encoder module; and (2) the accuracy can be fully recovered after finetuning, validating that our auto-encoder module helps to compress the Q/K vectors for reducing the amounts of costly data movements with negligible overheads and maintained model accuracy.

\subsection{The Unified ViTCoD Algorithm}
\label{sec:alg_unified}


\setlength{\columnsep}{10pt}%
\begin{wrapfigure}{r}{0.2\textwidth}
    \vspace{-1em}
    \centering
    \includegraphics[width=0.2\textwidth]{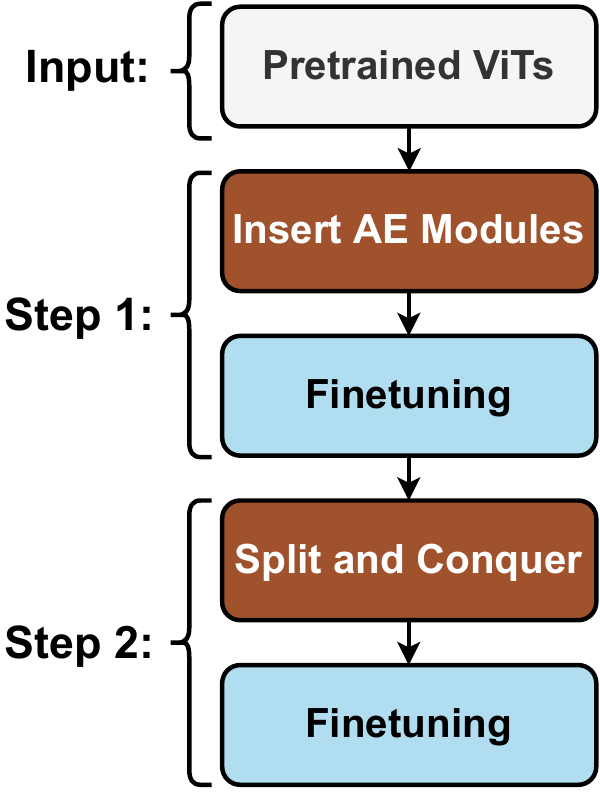}
    \vspace{-2em}
    \caption{The unified ViTCoD pipeline.}
    \vspace{-1em}
    \label{fig:vitcod_pipeline}
\end{wrapfigure}
The split and conquer algorithm and the auto-encoder module explore two orthogonal directions towards efficient ViT inference. Specifically, the former helps to reduce the number of attention computations and polarize imbalanced workloads to merely two patterns, while the latter explores the opportunity of trading costly data movements for cheaper computations to overcome the low PE utilization problem resulting from pruning unimportant attentions.
Our ViTCoD algorithm incorporates both into one unified pipeline. As shown in Fig. \ref{fig:vitcod_pipeline}, taking pretrained ViTs as inputs, we first insert AE modules to each attention head in Step 1, then conduct the split and conquer algorithm in Step 2, and finally perform finetuning to restore the model accuracy.
\hrhpca{We finetune both DeiT and LeViT for 100 epochs using the same training recipe as \cite{DeiT}, and finetune Strided Transformer for 10 epochs using the same training recipe as \cite{li2022exploiting}, except for adopting a smaller learning rate of 1e-5.}


\section{Proposed ViTCoD Accelerator}
\label{sec:method_accelerator}
\subsection{Motivation of ViTCoD Accelerator}
\label{sec:hardware_motivation}

\textbf{Opportunity 1: Fixed and Structurally Sparse Attention.}
Our ViTCoD split and conquer algorithm exhibits a great potential in both reducing the dominate attention computations and alleviating the irregularity of the resulting sparse attention masks.
However, this potential cannot be fully exploited by existing Transformer accelerators \cite{lu2021sanger,wang2021spatten,qu2022dota} due to the fact that (1) they are designed for dynamic sparse attention which requires both on-the-fly mask generation and highly reconfigurable architecture supports, both of which require nontrivial overheads, and (2) they are not dedicated for processing the enforced two distinct workloads, i.e., denser and sparser patterns, from our ViTCoD algorithm.
As such, our ViTCoD accelerator is motivated to exploit the new opportunities i.e., fixed and structurally sparse patterns, resulting from ViTCoD algorithm to boost ViTs' inference efficiency. 

\textbf{Opportunity 2: Compact $Q$ and $K$ Representation.}
Our ViTCoD auto-encoder (AE) module offers another opportunity to trade costly data movements for lower-cost computations, as it compresses both $Q$ and $K$ into a compact representations (i.e., 50\% of the origin size) for reduced data movements at the expense of a slightly increased computations for recovering them back for attention calculation.
This is an extremely effective enabler for boosting the PE utilization of sparse attention accelerators, where computing one attention score needs to load two complete $Q$ and $K$ vectors of large feature dimensions. Moreover, the loaded vectors are rarely reused due to both the high sparsity and diagonal pattern in the enforced sparser patterns as shown in Figs.~\ref{fig:split_and_conquer} and \ref{fig:attn_visual}, leaving data movements as a bottleneck and indispensably motivating our lightweight auto-encoder module.

\begin{figure}[t]
    \centering
    \includegraphics[width=\linewidth]{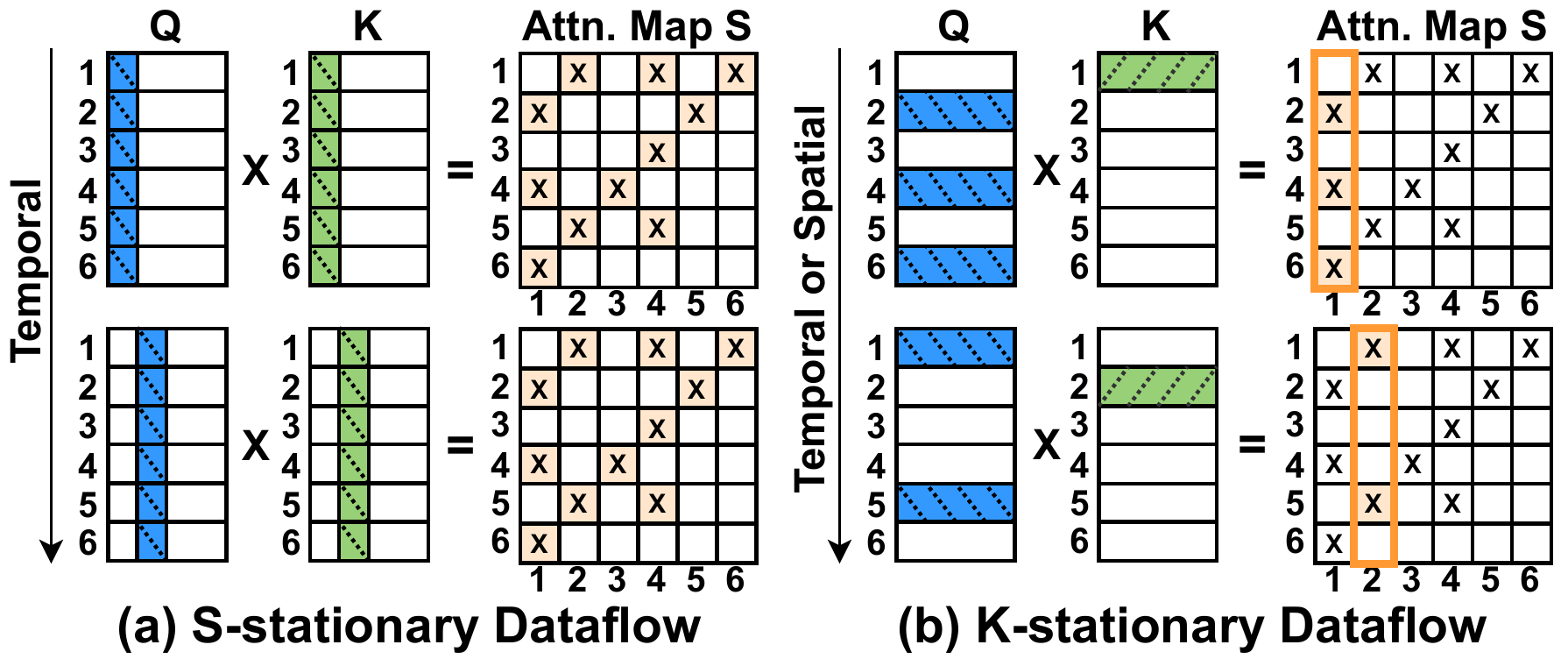}
    \vspace{-1.3em}
    \caption{Illustrating the $S$- and $K$-stationary dataflows.}
    \vspace{-2em}
    \label{fig:dataflows}
\end{figure}

\textbf{Design Exploration 1: Micro-architecture.}
For accelerating workloads of diverse sparse patterns, two typical designs can be considered: a single accelerator \cite{srivastava2020matraptor,qin2020sigma} or multiple sub-accelerators \cite{shen2017maximizing,zhang2018dnnbuilder} with each tailoring for one of the diverse computation patterns. 
The former simplifies the control logic but suffering from under-utilization if the sparsity of diverse patterns become high; the latter favors a high utilization yet can lead to a nontrivial control overhead for dealing with the diverse data dependency across sub-accelerators.      
Thanks to our proposed ViTCoD algorithm, 
\hry{there exist mostly two diverse workloads of either denser or sparser patterns as demonstrated in Sec. \ref{sec:alg_split_and_conquer}.}
As such, we consider the latter design with merely one denser engine and one sparser engine to efficiently process ViTCoD algorithm trained sparse ViTs for reducing both the scheduling and control overheads.
Additionally, our ViTCoD accelerator integrates encoder and decoder engines to support the AE modules.

\textbf{Design Exploration 2: Dataflows.}
Here we discuss two potential dataflows for processing the SDDMM of sparse attention modules, i.e., $S$- and $K$-stationary dataflows, and discuss their advantages and disadvantages.
As shown in Fig. \ref{fig:dataflows} (a), \textbf{$S$-stationary dataflow} executes different $Q$ and $K$ vectors in a parallel manner, where the features of each $Q/K$ are multiplied sequentially for accumulation.
Its advantage is that both $Q$ and $K$ vectors are fully reused after being loaded from an off-chip memory.
However, such $S$-stationary dataflow limits the acceleration efficiency from both the computation and storage aspects: (1) a low PE utilization for accelerating sparse attentions, because attention scores are spatially mapped in the PE array, where each PE corresponds to calculating one attention score, requiring high reconfigurablity and large control overhead to support sparse attentions; and (2) large on-chip register/buffer requirements to hold the intermediate partial sums of attention scores, which need to be stored on the registers of PE arrays for intra-PE accumulation.   
For example, Sanger \cite{lu2021sanger} adopts such a dataflow.
The other \textbf{$K$-stationary dataflow} is illustrated in Fig. \ref{fig:dataflows} (b), which loads $K$ vectors and multiplies them by different $Q$ vectors in a sequential manner, producing attention scores in a column by column manner, where feature dimensions of $Q/K$ are spatially mapped to the PE array for inter-PE accumulation.
Its advantage is that the $K$ vectors are fully reused and only a small on-chip buffer is needed for holding the intermediate results. In addition, it is more suitable for sparse attention acceleration since that only paired $Q/K$ vectors are multiplied according to the non-zero indexes in $S$, instead of spatially mapping and multiplying all $Q/K$ features as the $S$-stationary dataflow does.
Such benefits, however, come at the cost of requiring more frequent loading of  $Q$ vectors.
Given that our ViTCoD algorithm enables a high sparsity in attentions without hurting the model accuracy, and its trained ViTs result in partition between denser and sparser patterns along the $K$ dimension, the ideal dataflow should favor both sparse attention computations and supporting a dynamic number of $K$ vectors.
Therefore, the $K$-stationary dataflow is better suited for the sparse attention patterns resulting from our ViTCoD algorithm.
Moreover, the disadvantage of the $K$-stationary dataflow, i.e., more frequent $Q$ accesses, can be alleviated thanks to our AE modules, reducing the data movements.

\begin{figure*}[t]
    \centering
    \includegraphics[width=0.95\linewidth]{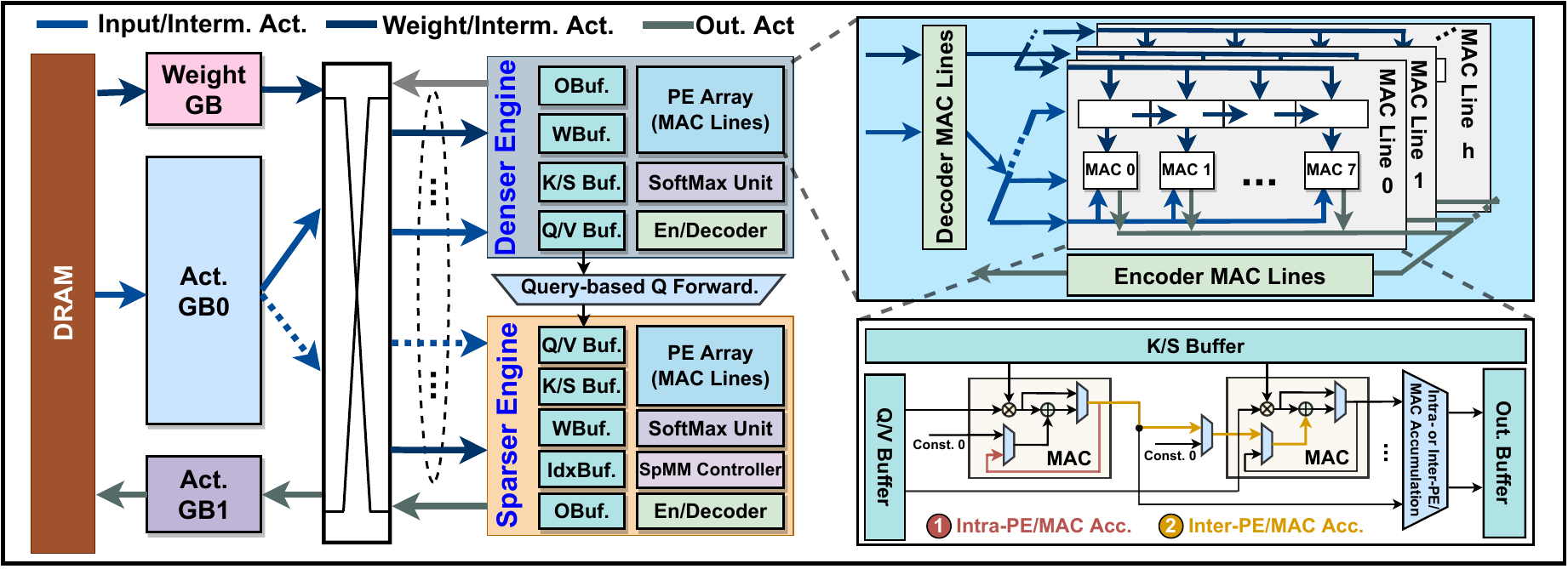}
    \vspace{-0.6em}
    \caption{Illustrating the micro-architecture of our ViTCoD accelerator.}
    \vspace{-1.3em}
    \label{fig:architecture}
\end{figure*}

\subsection{ViTCoD Accelerator's Micro-architecture}
\label{sec:micro-arch}
\vspace{-0.2em}


\textbf{Architecture Overview.}
\hry{Fig. \ref{fig:architecture} illustrates the overall micro-architecture of the proposed ViTCoD accelerator (\textbf{Left}: Memory hierarchy; \textbf{Middle}: Denser/Sparser engines; \textbf{Right}: PE array design).}
For efficiently processing the aforementioned two distinct denser and sparse workloads with reduced off-chip memory accesses and better PE/MAC utilization, our ViTCoD accelerator consists of two separate computing engines with each consisting of MAC lines and being dedicated to process the denser/sparser workloads of ViTCoD algorithm's resulting sparse attentions.
In addition, the accelerator integrates on-chip encoder and decoder engines to leverage our inserted AE modules in ViTs (See Sec. \ref{sec:alg_auto_encoder}) for much reduced data movements.
Specifically, \textbf{the Denser Engine} handles both the sampled dense $Q/K$ matrix multiplication during SDDMM (i.e., $Q \cdot K^T$) and the subsequent $S/V$ matrix multiplication during SpMM (i.e., $S \cdot V$).
%
\textbf{The Sparser Engine} at the same time handles the remaining irregular but significantly reduced workloads.
Such an engine is also capable of efficiently processing SpMM, SoftMax, and non-linear activation.
We term the proposed micro-architecture comprising two separate computing engines as \textit{two-pronged architecture} hereafter.
For minimal controlling overhead and interruption to the computing flow, these two engines are equipped with separate output buffers, so that their generated results can be written into the buffers in parallel.
Within the denser and sparser engines, we further design
\textbf{the Encoder/Decoder Engines} to leverage ViTCoD's AE module for trading costly data movements for lower-cost computations. 
Finally, our accelerator architecture can also be further reused for the $Q/K/V$ generation and MLP layers, where all MAC Lines are reconfigured to process these dense workloads.
Next, we elaborate the design details of each engine.

\subsubsection{Two-pronged Architecture}

\hry{\textbf{Denser Engine.}}
As shown in Sec. \ref{sec:alg_split_and_conquer}, the denser attention maps resulting from ViTCoD's split and conquer algorithm vary in terms of the number of global tokens among different layers/heads.
Thus, to balance the workload of processing the denser patterns of different attention heads, we adopt a dynamic PE allocation between the denser and sparser engines. Thanks to the fixed sparse attention masks known as a priori, we can easily estimate the workload size of the two patterns. As such, given the available hardware resource budgets, e.g., the number of PEs, the off-chip memory bandwidth, etc, we allocate hardware resource to each engine proportional to its assigned workload size.
Moreover, the denser engine also supports operations of MLPs, whose computation falls into general matrix multiplication (GEMM) that can be divided into multiple chunks and then processed in parallel.
Therefore, we dedicate each PE line (or MAC line) to the computation of one chunk. Note that here chunks are divided according to the attention heads, so that each line's data and intermediate results are unique to itself, avoiding unnecessary data movements as well as matrix split and concatenation among different PE lines.

\begin{figure}[t]
    \centering
    \includegraphics[width=0.9\linewidth]{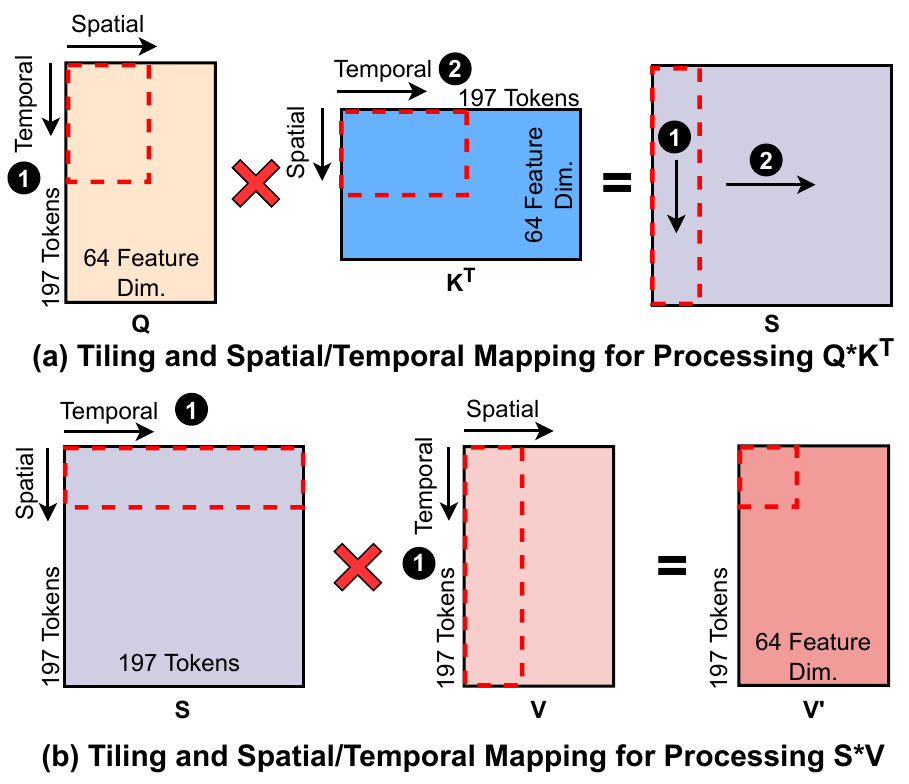}
    \vspace{-1em}
    \caption{Illustrating the tiling and spatial/temporal mappings for processing $Q \cdot K^T$ and $S \cdot V$.}
    \label{fig:tiling}
    \vspace{-1.8em}
\end{figure}

Since all attention heads are processed in parallel, the assigned PE lines for each head cannot afford the multiplication between $Q$ and $K$ vectors within one cycle. As such, we consider fine-grained tiling and carefully design the spatial/temporal mappings during both the SDDMM and SpMM phases.
As illustrated in Fig. \ref{fig:tiling}, for calculating $S=Q \cdot K^T$, we consider $K$-stationary dataflow, which favors our design as discussed in Sec. \ref{sec:hardware_motivation}. In particular, we tile the $Q/K$ vectors along the feature dimension and map them to the PEs spatially; then {\color{black}{\ding{182}}} multiply the loaded $K$ with all related $Q$ vectors temporally and accumulate their partial sums among different PEs, i.e., {\color{purple}{inter-PE accumulation}} \hry{\textbf{as shown in the lower-right part of Fig. \ref{fig:architecture}}} (See {\color{purple}{\ding{182}}}), to generate the first column of attention maps, after which {\color{black}{\ding{183}}} the next $K$ vector is loaded with similar mappings until all related attention scores are calculated.
For processing $V' = S \cdot V$, we consider output stationary instead to reduce the on-chip buffer requirements and to avoid frequently loading attention maps.
Specifically, we tile the $S/V$ vectors along the token dimension and map them to the PEs spatially; then {\color{black}{\ding{182}}} temporally accumulate the partial sums along the feature dimension for updating the $V$ vectors. Such a tiling and computation mapping fully reuses $S$ and $V$, and only requires a small on-chip buffer to hold the calculated outputs, for achieving such benefits, PE lines need to be reconfigured from {\color{purple}{inter-PE accumulation}} to {\color{olive}{intra-PE accumulation}} as shown in Fig. \ref{fig:architecture} (See {\color{olive}{\ding{183}}}).

\textbf{Sparser Engine.}
Benefiting from our ViTCoD algorithm, the sparser patterns feature a much reduced data and computation density (i.e., $>90\%$). As such, our ViTCoD accelerator handles this workload by utilizing (1) \textit{a CSC data format} for indexing the non-zeros in the sparser areas of the attention maps; and (2) \textit{query-based $Q$ forwarding}.
For (1), thanks to the drastically reduced density in the sparser areas, we are able to pre-load and store the indexes in a CSC data format, enabling the ViTCoD accelerator to load the required $Q/K$ more regularly. Also, we consider the CSC format instead of a Coordinate (COO) format for better matching with the adopted $K$-stationary dataflow, which produces attention maps column by column.
For (2), since the denser and sparser engines operate in parallel, when the sparser branch is working on a certain $Q$ vectors, it is likely that the denser engine is working on the same $Q$. Therefore, instead of directly loading them from the off-chip memory, we consider to first query the $Q$ buffer of the denser engine. Such a query is performed in an on-demand manner.
The tiling and spatial/temporal mappings of the sparser engine are consistent with that of the denser engine during both the SDDMM and SpMM phases excepts for that only non-zeros are calculated in the sparser engine thanks to the pre-stored indexes.
To support the computations and special operations of ViTs, both engines contain multiple functional units as described below.

\textbf{Architecture of the Denser/Sparser Engines.} Denser/Sparser \sloppy engines (see the middle of Fig. \ref{fig:architecture}) have the following functional units:
(1) \textit{Dedicated buffers} for outputs (\textbf{OBuf.}), weights (\textbf{WBuf.}), $K/S$ vectors (\textbf{K/S Buf.}), indexes (\textbf{IdxBuf.}), and $Q/V$ vectors (\textbf{Q/V Buf.}), each of them is equipped with parallel read/write ports to favor more local reuses, whose sizes are decided in the the resource allocation stage mentioned above;
(2) \textit{Sparse/Dense matrix multiplication controller} which supports both dense and sparse workloads. For the dense workload, it loads the corresponding two vectors and performs multiplication with either inter-PE or intra-PE accumulation; For the sparse workload, only non-zero elements and their indexes are loaded for calculation, thanks to the pre-loaded indexes in a CSC format;
(3) \textit{SoftMax units}
are used after a complete attention score is computed, we conduct an exponential operator for the softmax function following \cite{lu2021sanger};
and (4) \textit{Activation units} for the non-linear activation functions. Specifically, we use gating modules for ReLU and lookup tables to estimate other activation functions;
Additionally, we incorporate the encoder/decoder engines to cooperate with the inserted AE modules as described below. 

\subsubsection{Encoder and Decoder Engines}
\label{sec:encoder_and_decoder_engines}
Recalling that the inserted AE module proposed in our ViTCoD algorithm offers us a new opportunity to trade costly data movements for lower-cost computations.
To leverage such benefits, we design both the encoder and decoder engines in our ViTCoD accelerator, where the weights of the AE module are pre-loaded and stored on chip thanks to their small sizes (e.g., $6 \times 3$).
\hry{\textbf{As shown in the upper right part of Fig. \ref{fig:architecture},}} 
\hry{encoder and decoder have their own PE/MAC lines and are used to process the encoder/decoder workloads when needed, i.e., those PE/MAC lines will also be used to process other denser/sparser workloads when encode/decode are not needed.}
In particular, the encoder engine is enabled right after the linear projection for generating $Q/K/V$ so as to compress $Q/K$ before transferring them back to the off-chip memory. Also, their computation can be fully pipelined to hide the processing time of the encoder engine;
The decoder engine is then needed before loading $Q/K$ into the PE arrays, which will be pipelined with the data movements instead.

\begin{figure}[t]
    \centering
    \includegraphics[width=\linewidth]{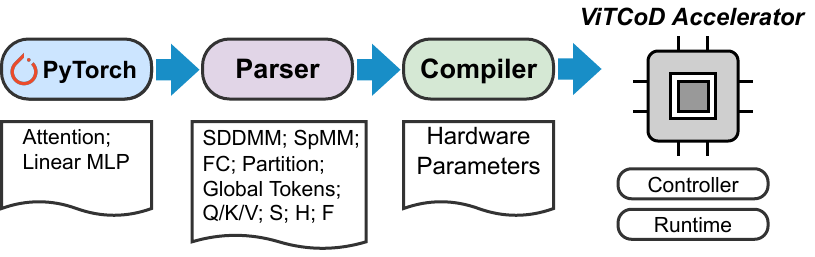}
    \vspace{-2em}
        \caption{The algorithm-hardware interface pipeline.}
    \label{fig:interface}
    \vspace{-1.5em}
\end{figure}




\subsubsection{Reconfigurability}
\label{sec:reconfigurability}

To support the potential need of task change after deployment, e.g., ViT models with different mask patterns or head numbers, the proposed ViTCoD accelerator is equipped with a low-cost hardware reconfigurable strategy for adaptation during the hardware compilation process, which only need one-time compilation cost for each task.
\hry{
Fig. \ref{fig:interface} shows the corresponding interface pipeline.
The given sparse ViT layers will first be passed through a network parser for extracting hardware configurations before feeding the hardware compiler to generate processing instructions.
According to the configurations extracted from ViTCoD algorithm, e.g., the number of global tokens, buffer sizes, and dataflows, the hardware compiler helps to generate corresponding instructions to enable ViTCoD accelerator to be reconfigured to reallocate on-chip memories and PEs/MACs in both the Denser and Sparser engines. The compiler generates instructions to control the ViTCoD accelerator to switch between the inter-PE/MAC accumulation (w.r.t. $K$-stationary dataflow during $Q \cdot K$) and intra-PE/MAC accumulation (w.r.t. output-stationary dataflow during $S \cdot V$) modes.
Moreover, the cost of such reconfigurability is amortized across the execution lifetime of each task.}

%% file: sections/4-Experiments.tex
\vspace{-0.2em}
\section{Experiments}
\subsection{Experiment Setting}
\label{sec:exp_setting}

\textbf{Models, Datasets, and Training Settings.}
\underline{Models:} We consider DeiT-Base/Small/Tiny~\cite{DeiT} which are well recognized ViT models, LeViT-128/192/256~\cite{graham2021levit} which are ViT variants targeting mobile devices, and Strided Transformer~\cite{li2022exploiting} which achieves SOTA performance on AR/VR applications.
\underline{Datasets:} We use ImageNet dataset~\cite{deng2009imagenet} for evaluating DeiT and LeViT on the image classification task, and Human3.6M dataset~\cite{ionescu2013human3} for evaluating Strided Transformer on the 3D human pose estimation task.
\underline{Training Settings:} We finetune both DeiT and LeViT using the same training recipe as \cite{DeiT}, and finetune Strided Transformer with the same recipe as \cite{li2022exploiting}, excepts for adopting a smaller learning rate of 1e-5.


\textbf{Baselines and Evaluation Metrics.}
\underline{Baselines:} To benchmark ViTCoD with SOTA attention accelerators, 
we consider a total of five baselines, including three general platforms: CPU (Intel Xeon Gold 6230R), 
\hrhpca{EdgeGPU (Nvidia Jetson Xavier NX),}
and GPU (Nvidia 2080Ti), and two attention accelerators: SpAtten~\cite{wang2021spatten} and Sanger~\cite{lu2021sanger}. 
Note that when benchmarking with GPUs w/ larger batch size, we scale up the accelerators' hardware resource to have a comparable peak throughput for a fair comparison following \cite{qu2022dota}.
\underline{Metrics:} We evaluate all platforms in terms of latency speedups and energy efficiency. In addition, we compared the achieved attention sparsity and model accuracy for all ViT models.

\begin{figure*}[t]
    \centering
    \includegraphics[width=0.96\linewidth]{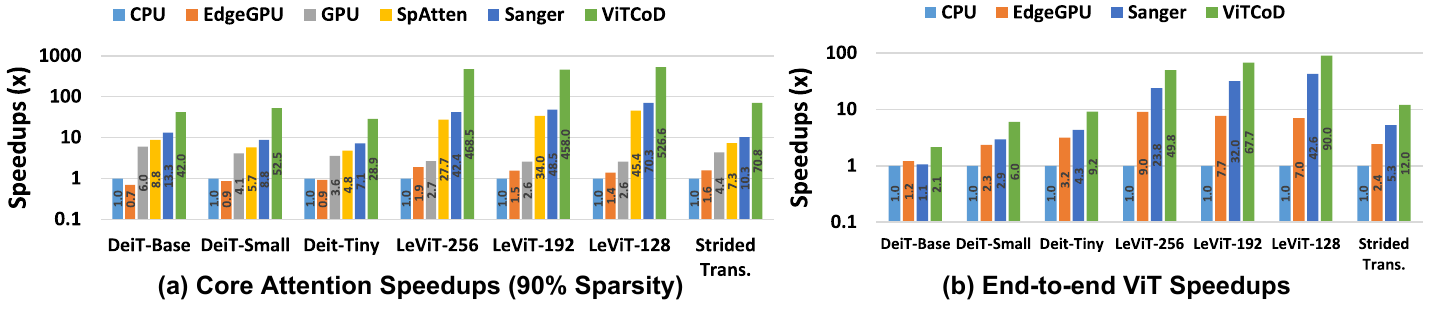}
    \vspace{-1.2em}
    \caption{The normalized speedups (w.r.t. CPU) achieved by ViTCoD over five SOTA baselines on Seven ViT models.}
    \label{fig:overall_comp}
    \vspace{-1em}
\end{figure*}

\begin{figure}[t]
    \centering
    \includegraphics[width=\linewidth]{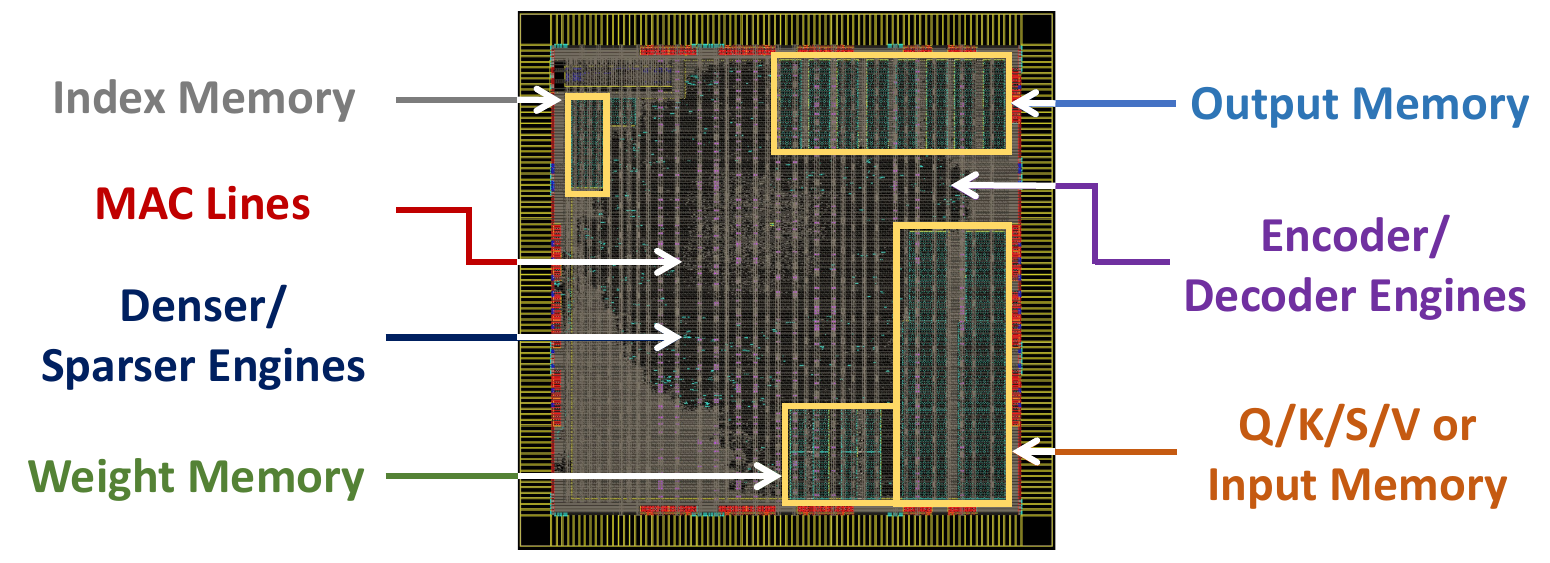}
    \vspace{-1.5em}
        \caption{Layout floorplan of our ViTCoD accelerator.}
    \label{fig:layout}
    \vspace{-1.5em}
\end{figure}

\textbf{Hardware Platform Setup.}
\underline{Characteristics:}
ViTCoD is designed with a total area of 3 $mm^2$, a DDR4-2400 (w/ multiple banks connected to one unified controller) memory bandwidth of 76.8GB/s, and power of 323.9mW at a core frequency of 500MHz, and is equipped with 320KB SRAM and 512 MACs (64 MAC lines with each having 8 MACs).
The SRAM includes (1) Act. GB0/GB1 of 256KB where $Q/K/S/V$ or input buffer occupies 128KB, index buffer occupies 20KB, and the output buffer occupies 108KB; and (2) Weight GB of 64KB.
\underline{Evaluation:} We develop a cycle-accurate simulator to simulate the performance of our ViTCoD accelerator, for which the MAC and memory access cost are derived from the post-layout simulation. 
Fig. \ref{fig:layout} shows its corresponding layout floorplan.
We verified it against the RTL implementation to ensure its correctness.
Specifically, we synthesize our RTL design with a commercial 28nm CMOS technology using Synopsys tools (Design Compiler for gate-level netlist~\cite{DC}; IC Compiler II for layout~\cite{ICCII}) and the Memory Compilers from the foundry.


\vspace{-0.2em}
\subsection{Overall Performance Comparison}
\label{sec:overall_comp}


Fig. \ref{fig:overall_comp} shows the overall performance of our ViTCoD and five baselines. 
We see that ViTCoD on-average achieves \hr{235.3$\times$, 160.6$\times$, and 86.0$\times$} core attention speedups over the general CPU, EdgeGPU, and GPU platforms, respectively.
Moreover, in terms of benchmarking on end-to-end ViT accelerations, ViTCoD on-average achieves \hr{33.8$\times$ and 5.6$\times$} over CPU and EdgeGPU, respectively.
We further compare the proposed ViTCoD with SOTA attention accelerators, SpAtten \cite{wang2021spatten} and Sanger \cite{lu2021sanger}, in terms of both core attention speedups and end-to-end ViT speedups.
Note that we implement and simulate both of them on ViTs with similar hardware configurations and areas for fair comparisons; Meanwhile, we test our simulators with their reported experiment results on NLP Transformers to ensure the correctness.
In particular, \textbf{for accelerating core attention workloads}, i.e., both SDDMM and SpMM phases, ViTCoD achieves \hr{10.1$\times$ and 6.8$\times$} speedups over SpAtten and Sanger, respectively, under 90\% sparsity of attention maps.
We also consider to compare them under 80\% sparsity, at which time ViTCoD achieves \hr{4.8$\times$ and 3.2$\times$} speedups over SpAtten and Sanger, respectively.
\textbf{For accelerating the end-to-end ViT models}, the speedups will be \hr{3.1$\times$ and 2.1$\times$}.
Note that here the end-to-end speedups are larger than 2$\times$ although self-attention only accounts for $>=50\%$ of the whole ViT model when executed on EdgeGPU, that is because 
we are comparing to other accelerators instead of ViTCoD's hardware w/o ViTCoD techniques. If we compare w/ ViTCoD and w/o ViTCoD, the speedups will be around 1.8$\times$ instead.
This set of comparisons validate the effectiveness and superiority of ViTCoD's dedicated algorithm and accelerator innovations: (1) the split and conquer algorithm for regular and fixed sparse attention maps and the corresponding two-pronged architecture design; (2) the AE module for less data movements and the related encoder/decoder engines.

\textbf{Discussion of NLP Models}.
For NLP models, ViTCoD algorithm’s enforced static sparse attention patterns can degrade the model accuracy, e.g., -1.18\% for 60\% sparsity vs. the unpruned counterparts for BERT-Base-NLP-models on the GLUE-MRPC dataset. For a fair comparison, we further consider the dynamic attention prediction overhead, under which ViTCoD's attention speedups is 1.93$\times$/3.69$\times$ for a sparsity of 60\%/90\% sparsity, respectively, over Sanger.

\vspace{-0.2em}
\subsection{Evaluation of the ViTCoD Algorithm}
\label{sec:alg_exps}

\begin{figure}[t]
    \centering
    \includegraphics[width=\linewidth]{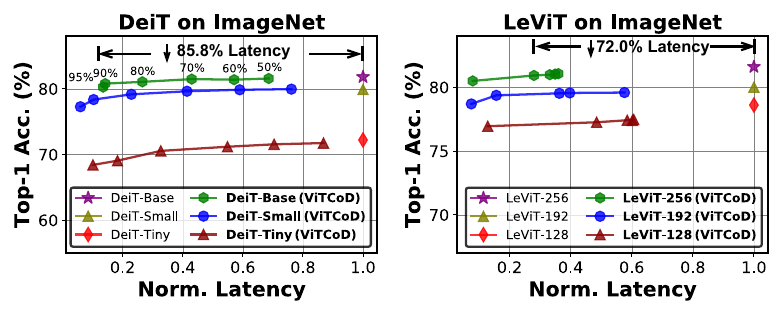}
    \vspace{-2.5em}
    \caption{Comparison between ViTCoD with unpruned baselines when adopting DeiT and LeViT on ImageNet.}
    \label{fig:comp_alg}
    \vspace{-1.em}
\end{figure}

As shown in Fig. \ref{fig:comp_alg}, we compare the accuracy and latency trade-offs of ViTCoD with unpruned baselines when evaluating the attention layer of six ViT models, i.e., DeiT-Base/Small/Tiny and LeViT-256/192/128, on ImageNet.
\hry{
Note that here we evaluate the complete ViTCoD algorithm with both split and conquer algorithm and inserted auto-encoder (50\% compression ratio, e.g., 12 heads $\rightarrow$ 6 heads).}
We observe that the split and conquer algorithm helps to reduce \hr{45.1\% $\sim$ 85.8\% and 72.0\% $\sim$ 84.3\%} latency of attention layers for DeiT and LeViT, respectively, while leading to a comparable model accuracy (i.e., $<$ 1\% accuracy drop).
\begin{figure}[t]
    \centering
    \includegraphics[width=\linewidth]{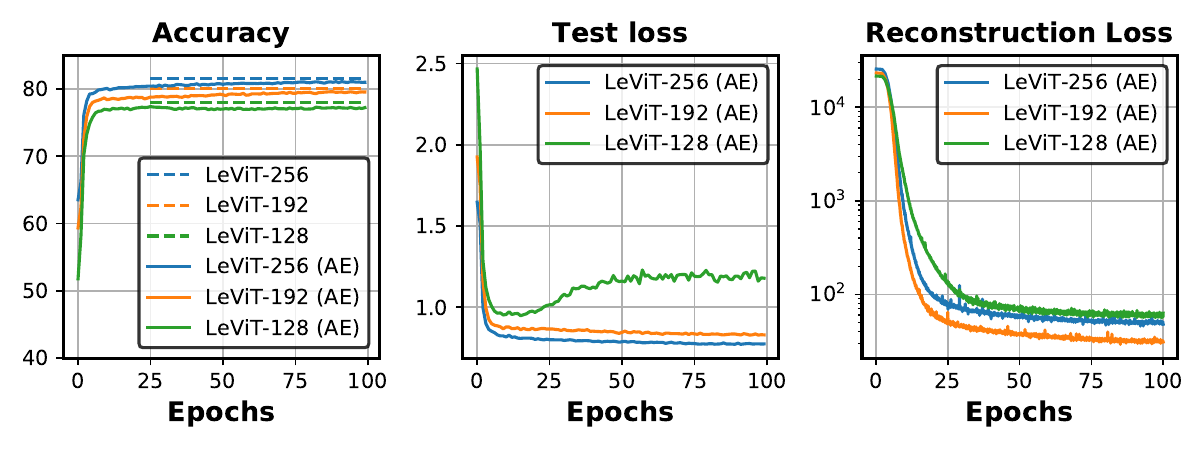}
    \vspace{-2.5em}
    \caption{Training trajectory of LeViTs with AE modules, where dashed lines denote vanilla LeViTs' accuracy.}
    \label{fig:ae_levit}
    \vspace{-1.5em}
\end{figure}
Also, we preform ablation studies of the split and conquer algorithm across a wide range of sparsity ratios, 
i.e., $\{50\%, 60\%, 70\%, 80\%, 90\%, 95\%\}$, and find that ViTCoD consistently achieves \hr{90\% and 80\%} sparsity on DeiT and LeViT models at a cost of negligible accuracy drop (i.e., $<$ 1\%).
\hry{
We also benchmark ViT models w/ and w/o the auto-encoder modules (w/o split and conquer algorithm), and show the training trajectory of DeiT models and LeViT models in Fig. \ref{fig:auto_encoder} (b) and Fig. \ref{fig:ae_levit}, respectively.
We observe that 
(1) both the test loss (i.e, $\mathcal{L}_{CE}$) and reconstruction loss are significantly reduced, verifying the convergence and effectiveness of our proposed ViTs incorporating learnable auto-encoder modules; and (2) the accuracy can be mostly recovered (with $<0.5\%$ accuracy drop) after finetuning, validating that our auto-encoder module helps to compress the Q/K vectors for reducing the amounts of costly data movements with negligible overheads and maintained model accuracy.
}
These experiments validate the effectiveness of ViTCoD's algorithm.

\hrhpca{
\textbf{Breakdown Pruning and Reordering.}
For quantitatively breakdown the benefits of pruning and reordering, we conduct experiments on DeiT-Base/Small/Tiny models.
Compared with reordering only, pruning makes sparse parts sparser, and thus enhances the polarization effect (i.e., more regular), offering on-average 5.14$\times$ speedups across 60\%/70\%/80\%/90\% pruning ratio (e.g., 8.14$\times$ speedups under 90\% sparsity).
Compared with pruning only, reordering makes the sparse pattern polarized and more regular, offering on-average 2.59$\times$ speedups across 60\%/70\%/80\%/90\% pruning ratios (e.g., 2.03$\times$ speedups under 90\% sparsity).
}

\begin{figure}[t]
    \centering
    \includegraphics[width=\linewidth]{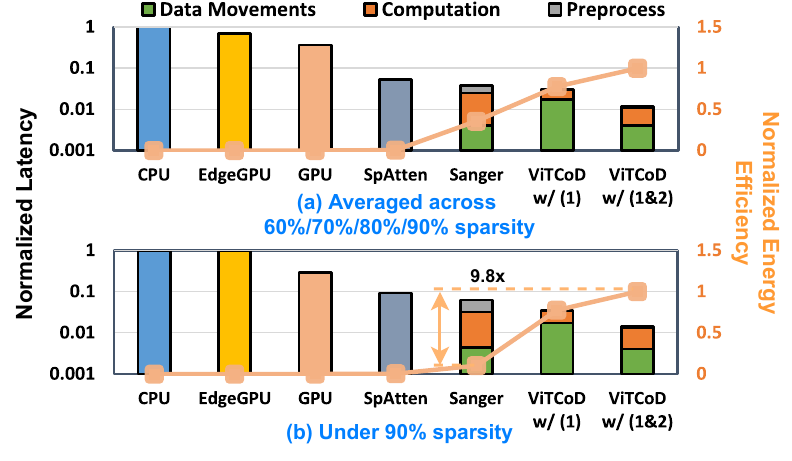}
    \vspace{-2em}
    \caption{\hrhpca{Comparison between ViTCoD and five baselines in terms of latency breakdown and energy efficiency.}}
    \vspace{-1.5em}
    \label{fig:comp_accelerator}
\end{figure}

\vspace{-0.3em}
\subsection{Evaluation of the ViTCoD Accelerator}
\label{sec:hardware_exps}

We first benchmark ViTCoD against five baselines. In Fig. \ref{fig:comp_accelerator}, ViTCoD consistently achieves both improved normalized efficiency and latency speedups over all baselines. Specifically, ViTCoD offers
\hrhpca{on-average (among six DeiT \& LeViT models mentioned in our experiment setting)} 
\hrhpca{235.3$\times$, 142.9$\times$, 86.0$\times$, 10.1$\times$, and 6.8$\times$}
attention speedups compared with CPU, EdgeGPU, GPU, SpAtten, and Sanger baselines when evaluating on attention layers of 90\% sparsity.
\hrhpca{For comprehensive evaluation, we evaluate across 60\%/70\%/80\%/90\% sparsity levels, the averaged speedups become
\hrhpca{127.2$\times$, 77.0$\times$, 46.5$\times$, 6.8$\times$, and 4.3$\times$} over CPU, EdgeGPU, GPU, SpAtten, and Sanger baselines, respectively.}
Meanwhile, ViTCoD maintains a high energy efficiency, achieving \hr{9.8$\times$} improvement over the most competitive baseline Sanger \cite{lu2021sanger}. We attribute its benefits to two explored opportunities for sparse ViTs: (1) fixed and structurally sparse attention maps and (2) compact $Q$ and $K$ representation, from both algorithm and hardware aspects.

\textbf{Latency Breakdown Analysis.}
To further evaluate and quantify the impact of each ViTCoD's innovation, we provide the latency breakdown analysis of both Sanger and ours ViTCoD as follows.
First, we separate the benefits of two innovations: (1) the split and conquer algorithm; and (2) the auto-encoder module.
\hry{As shown in Fig. \ref{fig:comp_accelerator}, ViTCoD with (1) leads to on-average 2.7$\times$ speedups over the most competitive baseline Sanger \cite{lu2021sanger}, on top of which adopting (2) further leads to 2.5$\times$ speedups.}
Moreover, we provide detailed breakdown in terms of computation, preprocess, and data movements.
Note that here data movements mean the overlapped computations and data movements.
We can see that:
(1) Sanger's adopted $S$-stationary dataflow fully reuses the loaded $Q/K$ vectors, reducing data movements at the cost of large computation workloads;
(2) ViTCoD's data movements are largely reduced from \hr{50\% to 28\%} after adopting AE modules, indicating the effectiveness of ViTCoD in alleviating the performance bottleneck.

%% file: sections/5-Conclusion.tex
\vspace{-0.3em}
\section{Conclusions}
\vspace{-0.1em}

We present ViTCoD, the first algorithm and accelerator co-design framework for sparse ViTs.
On the algorithm level, ViTCoD integrates (1) a split and conquer algorithm to prune and polarize the attention maps to be either denser or sparser with fixed masks and (2) an auto-encoder module to trade costly data movements for cheaper computations, without compromising the model accuracy.
On the hardware level, ViTCoD incorporates (1) a dedicated two-pronged accelerator to process each of the aforementioned denser or sparser workloads and (2) encoder and decoder engines to cooperate with auto-encoder modules, boosting the overall utilization and acceleration efficiency.
Extensive experiments consistently validate the advantages of ViTCoD over other accelerators.